\documentclass[10pt,journal,compsoc,standalone]{IEEEtran}
\ifCLASSOPTIONcompsoc
  \usepackage[nocompress]{cite}
\else
  \usepackage{cite}
\fi
\usepackage{amssymb}
\usepackage{xcolor} 
\usepackage{amsmath}
\usepackage{microtype}
\usepackage{graphicx}
\usepackage{multirow}
\usepackage{subfigure}
\usepackage{stfloats}
\usepackage{booktabs}
\usepackage{setspace}
\usepackage[breaklinks=true,bookmarks=false,colorlinks]{hyperref}
\usepackage{algorithm}
\usepackage{algorithmicx,algpseudocode}
\usepackage{xspace}
\usepackage{wrapfig}
\usepackage{lipsum}
\usepackage{enumitem}
\usepackage{diagbox}
\usepackage{cite}
\usepackage{listings}
\usepackage{colortbl}
\lstdefinestyle{mystyle}{
    basicstyle=\ttfamily,
    backgroundcolor=\color{white},   
    commentstyle=\color{green},      
    keywordstyle=\color{blue},       
    stringstyle=\color{red},         
    basicstyle=\ttfamily\footnotesize,
    breakatwhitespace=false,       
    breaklines=true,                
    showspaces=false,                
    showstringspaces=false,         
    showtabs=false,                  
    tabsize=4,                        
    frame=single,                   
    framesep=1pt,                   
    framerule=0.8pt,                 
    rulecolor=\color{black},        
}

\usepackage{makecell}
\usepackage{tikz}
\usepackage{forest}
\usetikzlibrary{trees,positioning,shapes,shadows,arrows.meta}
\definecolor{LightGray}{gray}{0.95}

\definecolor{MyDarkGreen}{rgb}{0.02,0.6,0.02}

\begin{document}
\title{Deep Time Series Models:\\ A Comprehensive Survey and Benchmark}

\author{Yuxuan~Wang,
        Haixu~Wu,
        Jiaxiang~Dong,
        Yong~Liu,
        Chen~Wang,
        Mingsheng~Long,
        Jianmin~Wang
\IEEEcompsocitemizethanks{
\IEEEcompsocthanksitem The authors are with the School of Software, BNRist, Tsinghua University, Beijing 100084, China. E-mail: wangyuxu22@mails.tsinghua.edu.cn.
\IEEEcompsocthanksitem Yuxuan Wang, Haixu Wu and Jiaxiang Dong contributed equally.
\IEEEcompsocthanksitem Corresponding authors: Mingsheng Long, Jianmin Wang.}
}

\markboth{IEEE Transactions on Pattern Analysis and Machine Intelligence,~Vol.~XX, No.~X}%
{Wang \MakeLowercase{\textit{et al.}}: Deep Time Series Models: A Comprehensive Survey and Benchmark}

\IEEEtitleabstractindextext{%
\begin{abstract}

Time series, characterized by a sequence of data points organized in a discrete-time order, are ubiquitous in real-world scenarios. Unlike other data modalities, time series present unique challenges in learning and modeling due to their intricate and dynamic nature, including the entanglement of nonlinear patterns and time-variant trends. Recent years have witnessed remarkable breakthroughs in time series analysis, with techniques shifting from traditional statistical methods to contemporary deep learning models. In this paper, we delve into the design of deep time series models across various analysis tasks and review the existing literature from two perspectives: basic modules and model architectures. Further, we develop and release Time Series Library (TSLib) as a fair benchmark of deep time series models for diverse analysis tasks. TSLib implements 41 prominent models, including both small- and large-scale time series models, covers 30 datasets from different domains, and supports 5 prevalent analysis tasks. Based on TSLib, we evaluate 16 popular deep time series models and 6 advanced time series foundation models. Empirical findings indicate that models with specific structures are apt only at distinct analytical tasks, providing insights for research and adoption of deep time series models. Code and datasets are available at \href{https://github.com/thuml/Time-Series-Library}{https://github.com/thuml/Time-Series-Library}.
\end{abstract}

\begin{IEEEkeywords}
Time series analysis, Deep time series models, Time series foundation models, Survey, Benchmark
\end{IEEEkeywords}}

\maketitle

\section{Introduction}
\IEEEPARstart{T}{ime} series refers to a sequence of data points indexed in a discrete-time order \cite{box2015time,hamilton2020time}, which are omnipresent in real-world applications, such as financial risk assessment, energy sustainability, and weather forecasting. Driven by the increasing availability of vast amounts of time series data across various domains, the community of time series analysis has witnessed tremendous advancements. Compared to image and text data, which have objectively prescribed syntax or intuitive patterns, the semantic information of time series data is primarily derived from the temporal variation~\cite{wu2023timesnet}. This presents significant challenges in understanding such data, notably in identifying sequential dependencies, trends, seasonal patterns, and complicated dynamics. Consequently, analyzing time series data necessitates sophisticated methods to extract its meaningful temporal representations.

Time series analysis has long stood as a pivotal research area, owing to its indispensable role in real-world applications \cite{jin2021trafficbert,xu2022anomaly,wu2023interpretable}. It involves scrutinizing temporal variations in sequential data to uncover underlying patterns, thereby facilitating accurate predictions and informed decision-making. A fundamental aspect of this process lies in identifying intricate temporal dependencies and variable correlations inherent in the data. By capturing these entangled dependencies, time series models reveal underlying dynamics and support a broad spectrum of downstream tasks, including forecasting, classification, imputation, and anomaly detection.

Traditional approaches, such as Autoregressive Integrated Moving Average (ARIMA) \cite{box2015time}, Exponential Smoothing, and Spectral Analysis \cite{koopmans1995spectral}, have long served as stalwart tools for analyzing time series data. Grounded in statistical methodologies, these methods have been instrumental in uncovering patterns, trends, and seasonality within temporal variations. However, their rigid assumptions of linearity and stationarity limit their application to dynamic and evolving data flows, making them inadequate for capturing the intricate nonlinear relationships and long-term dependencies present in real-world time series data.

\begin{table*}[tb]
  \setlength{\abovecaptionskip}{0cm}
  \setlength{\belowcaptionskip}{0cm}
  \vspace{-5pt}
\caption{Difference between our work and other related surveys. We present a systematic review of tasks and models, and offer an extensive benchmark.}
  \label{tab:SueveyCompare}
  \centering
  \begin{small}
  \renewcommand{\multirowsetup}{\centering}
  \setlength{\tabcolsep}{3.2pt}
  \renewcommand\arraystretch{1.1}
  \begin{tabular}{l|cccccccccc}
\toprule
\multirow{2}{*}{\scalebox{0.9}{Survey}} & \multicolumn{4}{c}{\scalebox{0.9}{Analysis Task}} & \multicolumn{5}{c}{\scalebox{0.9}{Model Architecture}}    & \multirow{2}{*}{\scalebox{0.9}{Benchmark}}  \\ 
\cmidrule(lr){2-5}\cmidrule(l){6-10}
                        & \scalebox{0.9}{Forecasting} & \scalebox{0.9}{Classification} & \scalebox{0.9}{Imputation} & \scalebox{0.9}{Anomaly Detection} &  \scalebox{0.9}{MLP} & \scalebox{0.9}{CNN} & \scalebox{0.9}{RNN} & \scalebox{0.9}{GNN} & \scalebox{0.9}{Transformer} &   \\ 
\midrule
\scalebox{0.9}{\cellcolor{LightGray}Fawaz \textit{et al.} (2019) \cite{ismail2019deep}} & \cellcolor{LightGray} & \cellcolor{LightGray}\checkmark  & \cellcolor{LightGray}  & \cellcolor{LightGray}    &  \cellcolor{LightGray}\checkmark & \cellcolor{LightGray}\checkmark   & \cellcolor{LightGray}\checkmark & \cellcolor{LightGray} & \cellcolor{LightGray} & \cellcolor{LightGray} \\
\scalebox{0.9}{Braei \textit{et al.} (2020) \cite{braei2020anomaly}} &  &  &   & \checkmark  &  \checkmark & \checkmark   & \checkmark & &  \\
\scalebox{0.9}{\cellcolor{LightGray}Torres \textit{et al.} (2021)} \cite{torres2021deep} & \cellcolor{LightGray}\checkmark  & \cellcolor{LightGray} &  \cellcolor{LightGray} &  \cellcolor{LightGray}  &  \cellcolor{LightGray}  & \cellcolor{LightGray}\checkmark & \cellcolor{LightGray}\checkmark   & \cellcolor{LightGray}\checkmark & \cellcolor{LightGray} & \cellcolor{LightGray} \\
\scalebox{0.9}{García \textit{et al.} (2021)} \cite{blazquez2021review} &  &  &   & \checkmark  &    & \checkmark & \checkmark   & \checkmark & &  \\
\scalebox{0.9}{\cellcolor{LightGray}Wen \textit{et al.} (2022) \cite{wen2022transformers}} & \cellcolor{LightGray}\checkmark & \cellcolor{LightGray}\checkmark  &  \cellcolor{LightGray} & \cellcolor{LightGray}\checkmark  &  \cellcolor{LightGray}  & \cellcolor{LightGray} & \cellcolor{LightGray}  & \cellcolor{LightGray} &\cellcolor{LightGray}\checkmark & \cellcolor{LightGray} \\
\scalebox{0.9}{Jin \textit{et al.} (2023) \cite{jin2023survey}} & \checkmark &\checkmark  & \checkmark   & \checkmark  &    &  &   & \checkmark 
 & & \\
\scalebox{0.9}{\cellcolor{LightGray}Shao \textit{et al.} (2023) \cite{shao2023exploring}} & \cellcolor{LightGray}\checkmark & \cellcolor{LightGray}  &  \cellcolor{LightGray}   &   \cellcolor{LightGray} & \cellcolor{LightGray}\checkmark   & \cellcolor{LightGray}\checkmark & \cellcolor{LightGray}\checkmark  &  \cellcolor{LightGray}\checkmark & \cellcolor{LightGray}\checkmark & \cellcolor{LightGray}\checkmark \\
\scalebox{0.9}{Qiu \textit{et al.} (2024) \cite{qiu2024tfb}} & \checkmark &   &     &    & \checkmark   & \checkmark & \checkmark  &  \checkmark & \checkmark & \checkmark \\
\midrule
\scalebox{0.9}{\cellcolor{LightGray}This Survey} &  \cellcolor{LightGray}\checkmark & \cellcolor{LightGray}\checkmark & \cellcolor{LightGray}\checkmark & \cellcolor{LightGray}\checkmark    & \cellcolor{LightGray}\checkmark & \cellcolor{LightGray}\checkmark & \cellcolor{LightGray}\checkmark    & \cellcolor{LightGray}\checkmark            &  \cellcolor{LightGray}\checkmark & \cellcolor{LightGray}\checkmark          \\
\bottomrule
  \end{tabular}
  \end{small}
  \vspace{-10pt}
\end{table*}

Deep models have garnered wide attention and achieved remarkable performance across various domains, including natural language processing (NLP) \cite{devlin2018bert}, \cite{achiam2023gpt}, computer vision (CV) \cite{he2016deep}, \cite{dosovitskiy2020vit}, and recommender systems~\cite{zhang2019deep}. 
In recent years, deep learning models \cite{zhou2021informer,wu2021autoformer, Yuqietal-2023-PatchTST,oreshkin2019nbeats,wu2023timesnet} have proven highly effective in capturing the intricate dependencies within time series data, making them a powerful tool for time series analysis over traditional statistical methods. 
More recently, Transformer models with attention mechanisms, originally developed for natural language processing tasks, have exhibited unprecedented capability in processing large-scale data \cite{achiam2023gpt} and have also been adapted for learning time series data. These architectures offer the advantage of selectively focusing on different parts of the input sequence, allowing for more nuanced discovery of both temporal and variable dependencies in time series.

While various time series models designed for different analysis tasks have emerged in recent years, there is a lack of a comprehensive overview of existing methods, covering both tasks and models. Previous reviews focus exclusively on either a specific model architecture or an analysis task. For example, \cite{ismail2019deep, braei2020anomaly, torres2021deep, blazquez2021review} each reviews deep learning methods for specific time series analysis tasks but fails to include advanced architecture such as Transformers. Several surveys \cite{wen2022transformers, jin2023survey} provide up-to-date reviews for time series analysis focusing on specific deep learning architectures (e.g., Graph Neural Networks and Transformers). Recently, BasicTS \cite{shao2023exploring} and TFB \cite{qiu2024tfb} introduce forecasting benchmarks that enable an unbiased evaluation of existing approaches but do not provide an overview of the architectural design of those deep models.

In this survey, we try to provide a unique review of deep time series models for researchers and practitioners, starting from the basic modules to modern architectures. To foster practical applications, a time series benchmark is offered for a fair evaluation and to identify the effective scope of existing models. This survey is organized as follows. Section~\ref{sec:basic} provides the background concepts of time series analysis. Section~\ref{sec:module} introduces the basic modules that are widely utilized in prevalent deep time series models. Section~\ref{sec:arch} reviews existing deep time series models in terms of the architecture design. Section~\ref{sec:ltsm} discusses recent processes in time series foundation models. Section~\ref{sec:tslib} introduces an easy-to-use open-source benchmark, Time Series Library (TSLib), along with extensive experimental comparisons and detailed analyses. Section~\ref{sec:future} provides a brief discussion of future research directions while Section~\ref{sec:conclusion} summarizes this survey.

\section{Preliminaries}\label{sec:basic}

\subsection{Time Series}

Time series is a sequence of $T$ observations ordered by time, which can be denoted as $\mathbf{X} = \{\mathbf{x}_1, \mathbf{x}_2, \cdots, \mathbf{x}_T\} \in \mathbb{R}^{T\times C}$, where $\mathbf{x}_t \in \mathbb{R}^{C}$ represents the observed values at timestamp $t$ and $C$ is the number of variables. Note that spatiotemporal data (e.g.~videos or weather records) can also be viewed as a special type of time series, where the ``variate dimension'' of the above notion corresponds to spatial pixels \cite{van2016pixel} in a video. However, previous research on spatiotemporal data primarily relies on highly structured spatial information~\cite{oprea2020review}, which limits their applications to general variates, like heterogeneous variates with distinct semantic meanings. In this paper, we focus on the more generic formalization, where we do not assume any spatial structure inside the variate dimension. Under this definition, the essence of time series analysis is to capture and utilize the temporal dependencies and inter-variable correlations within the observations.

\textbf{Temporal Dependency} Given the inherent sequential nature of time series data, one evident technological paradigm is to capture the temporal dependence of a set of historical data. Essentially, temporal dependencies refer to the intricate connections that manifest among distinct time points or sub-series. Traditional statistical methods, such as those based on autoregressive principles represented by ARIMA \cite{box2015time}, have been widely studied to model these dependencies. Owing to their simplicity and interpretability, these statistical methods retain prominence in scenarios where the underlying temporal dynamics exhibit limited complexity. 

\textbf{Variate Correlation} Beyond temporal dependencies, comprehending the correlations among distinct variables is paramount for analyzing multivariate time series, particularly in high-dimensional contexts. These correlations encapsulate the complex interactions and associations among multiple measurements as they evolve over time. Elucidating these relationships provides valuable insights into the underlying dynamics and latent processes. Conventional approaches, exemplified by VAR \cite{stock2001vector}, extend the concept of autoregression to multiple variables, thereby enabling the capture of inter-variable relationships across time. However, VAR models represent each variable as a linear combination of lagged values of all other variables, which inherently limits their ability to capture misaligned or nonlinear relationships.

The aforementioned fundamental elements of time series present considerable challenges for constructing practical analysis models. While traditional statistical methods offer foundational theoretical insights, their inherent reliance on presumptions of the data, such as linearity and stationarity, often restricts their applicability in intricate real-world scenarios.  Recently, deep learning models have emerged as powerful alternatives for time series modeling. Their capacity for representation learning and scalability to big data uniquely positions them for high-dimensional and nonlinear time series data, effectively fulfilling diverse analysis tasks.

\subsection{Time Series Analysis Tasks}

\begin{figure}[t]
    \centering
    \includegraphics[width=0.95\linewidth]{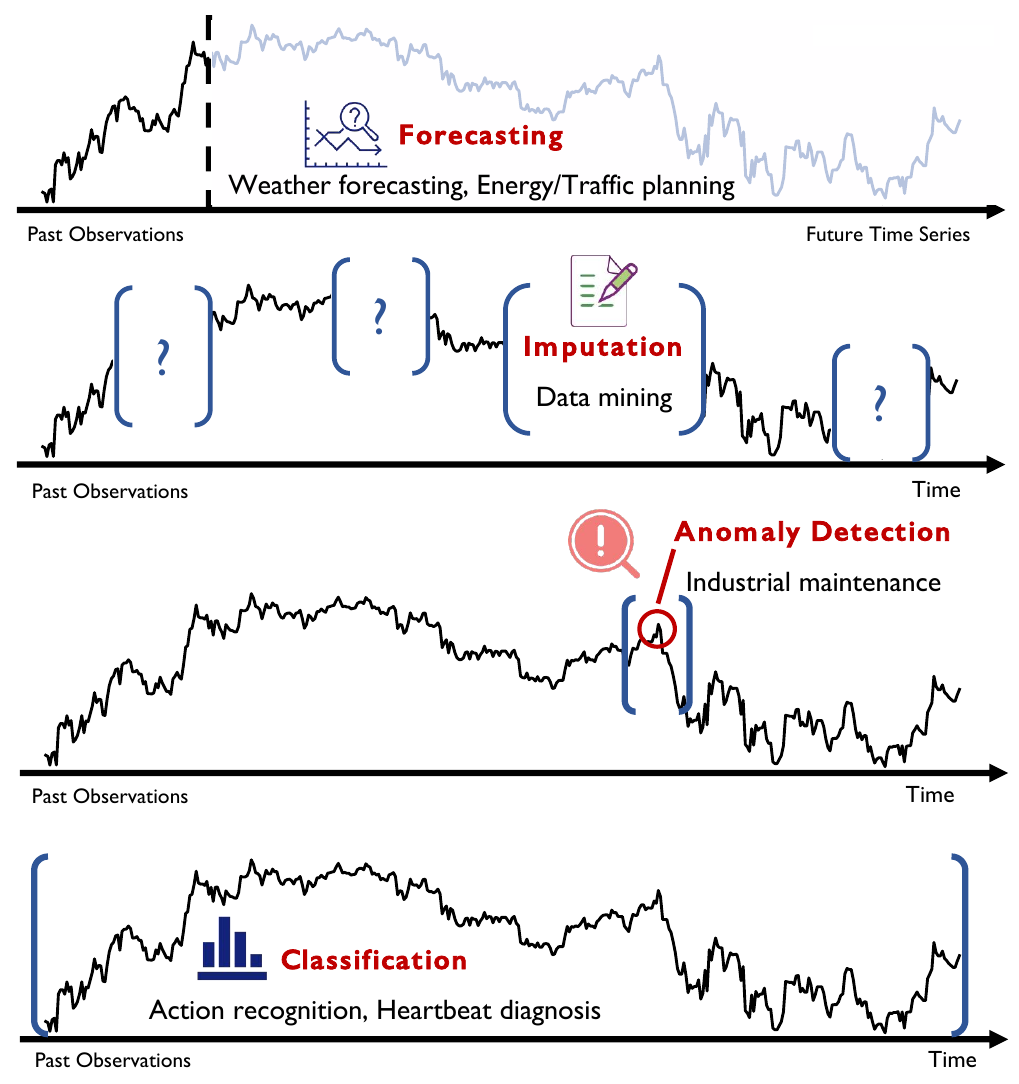}
    \vspace{-5pt}
    \caption{Schematic illustration of different time series analysis tasks.}
    \label{fig:analysis-task}
    \vspace{-5pt}
\end{figure}

Based on underlying patterns and trends within time series data, time series analysis encompasses various downstream applications, including forecasting \cite{weigend2018time,zhang2003time}, imputation \cite{fang2020time, luo2018multivariate, moritz2017imputets}, classification \cite{ismail2019deep, zhao2017convolutional}, and anomaly detection \cite{laptev2015generic,xu2022anomaly}, each finding distinct applications in diverse domains, as illustrated in Figure \ref{fig:analysis-task}. 

Forecasting is a fundamental task in time series analysis that requires models to uncover temporal dependencies and dynamic patterns. By capturing the relationships between past and future data, the forecasting model aims to predict future values or trends of the input series. Missing data due to sensor failures, data corruption, or absent measurements is ubiquitous in practical applications, leading to a growing demand for time series imputation to obtain higher-quality data. Unlike forecasting, which predicts future values based on historical observations, imputation focuses on reconstructing missing values using the available contextual information. Anomaly detection involves identifying unusual or abnormal patterns within a time series, which can indicate critical events, system faults, or outliers requiring further investigation. Lastly, classification assigns a label or category to a given time series based on its characteristics, a task widely utilized in fields such as medical diagnosis.

Given the pluralism of time series analysis, an array of deep learning models have emerged to address specific challenges in different tasks. Instead of a task-centric view into exhaustive details, this survey takes a model-centric view of deep time series models, focusing on their basic modules and architectural designs for addressing the foundational problems of capturing both temporal and inter-variable dependencies underlying large-scale time series datasets.

\section{Basic Modules}\label{sec:module}
In the past decade, time series modeling approaches have evolved significantly, shifting from traditional statistical models to sophisticated deep learning models. Despite these advancements, many classical tools and analytical algorithms remain widely used and continue to serve as foundational design principles in modern deep models. In this section, we focus on the major tools of classical time series analysis and demonstrate how they have been integrated as fundamental components in contemporary deep time series models.

\subsection{Stationarization}
Stationarity is a foundational concept in time series analysis, referring to the property where the statistical properties of a time series remain constant over time. A stationary time series exhibits a constant mean and variance, which simplifies statistical analysis and makes it easier to capture the underlying patterns and behaviors within a time series. Given that many statistics-based time series analysis methods fundamentally presuppose stationarity, the transformation of non-stationary data into a stationary form has become an essential preprocessing step. Conventionally, this stationarization is achieved through methods such as differencing or log-transformation. In recent deep learning approaches, data normalization \cite{ulyanov2016instance} takes the role of stationarization in a simple but effective way, which standardizes the value distribution of observations while maintaining the intrinsic variations and further helps mitigate the distribution shift between the source and target domains.

Recent deep learning architectures have developed specialized layers to address the non-stationary problem. The deep adaptive input normalization (DAIN) layer~\cite{passalis2019deep} was proposed to adaptively standardize time series data based on its original distribution. RevIN \cite{kim2021reversible} introduces a reversible instance normalization technique with learnable affine transforms to make the model bypass the non-stationary inputs.
Further, Non-Stationary Transformer \cite{liu2022non} (\emph{Stationary} for short in the following) proposes a simpler but more effective series stationarization technique that improves the predictive capability of non-stationary series without extra parameters. Specifically, for a sequence with $T$ time stamps and $C$ variates $\mathbf{X} = \{\mathbf{x}_1, \mathbf{x}_2, \cdots, \mathbf{x}_T\} \in \mathbb{R}^{T\times C}$, where $\mathbf{x}_t \in \mathbb{R}^{C}$ , the outline of Stationary \cite{liu2022non} can be summarized as follows:
\begin{equation}
\begin{aligned}
    \mu_\mathbf{x} &= \frac{1}{T} \sum_{i=1}^T \mathbf{x}_i,\ \sigma_\mathbf{x}^2 = \frac{1}{T} \sum_{i=1}^T (\mathbf{x}_i - \mu_\mathbf{x})^2, \\
    {\bar{\mathbf{X}}} &= \frac{\mathbf{X} - \mu_\mathbf{x}}{\sqrt{\sigma_\mathbf{x}^2 + \epsilon}}, \ {\bar{\mathbf{Y}}}=\operatorname{Model}({\bar{\mathbf{X}}}), \
    \hat{\mathbf{Y}} = \sigma_\mathbf{x}{\bar{\mathbf{Y}}} + \mu_\mathbf{x},
\end{aligned}
\label{equ: norm}
\end{equation}

where $\epsilon$ is a small value for numerical stability. $\mu_\mathbf{x},\sigma_\mathbf{x}^2\in\mathbb{R}^{1\times C}$ are the variate-specific mean and variance. To recover the distribution and non-stationarity of the original series, a denormalization module is further used to augment the model output ${\bar{\mathbf{Y}}}$ with mean and variance statistics of inputs.

The idea of stationarization and the aforementioned techniques have seen widespread adoption in subsequent deep time series models \cite{Yuqietal-2023-PatchTST, liu2023itransformer, gao2023client}. For instance, recent works such as SAN \cite{liu2024adaptive} rethink the nature of non-stationary data by normalizing non-overlapping, equally-sized slices instead of the entire series, and incorporating a statistics prediction module to forecast future slice distributions.

\subsection{Decomposition}

Decomposition \cite{cleveland1990stl, anderson1976time}, as a conventional approach in time series analysis, breaks time series into several components of more regular patterns, with a divide-and-conquer paradigm to explore complex series variations. In the previous work, a set of typical decomposition schemes are explored.

\subsubsection{Seasonal-Trend Decomposition}

Seasonal-trend decomposition \cite{rb1990stl, dagum2016seasonal} is one of the most common practices to enhance the predictability of raw time series data. This technique separates a series into distinct additive components: trend, seasonal, cyclical, and irregular.  The trend component represents the overall long-term pattern of the data over time, the cyclical component reflects repeated but non-periodic fluctuations within the data, the seasonal component indicates the repetitive patterns over a fixed period, and the irregular component is the residual or remainder of the time series after the other components have been removed. 

Such decomposition can be achieved using mathematical tools like filters or exponential smoothing \cite{wen2019robuststl, de2011forecasting}. Previous statistical approaches mainly adopt the trend-seasonality decomposition as data pre-processing \cite{taylor2018forecasting}.  In deep models, Autoformer \cite{wu2021autoformer} firstly introduces the idea of decomposition to deep learning architecture and proposes a \emph{series} \emph{decomposition} \emph{block} as a basic module based on the average pooling to extract the seasonal and trend-cyclical parts of input series or deep features:
\begin{equation}
\begin{aligned}
      \mathbf{X}_\mathcal{T} &= \operatorname{AvgPool}\left(\operatorname{Padding}(\mathbf{X})\right), \\
    \mathbf{X}_\mathcal{S} &= \mathbf{X} - \mathbf{X}_\mathcal{T}.
\end{aligned}
\end{equation}
Implemented based on a temporally average pooling layer, the proposed decomposition block effectively captures the trends $\mathbf{X}_\mathcal{T}$, and the remainder is taken as the seasonal part $\mathbf{X}_\mathcal{S}$. This design has been widely adopted as a native building block in the subsequent works \cite{wu2023interpretable, zeng2023transformers, cao2023inparformer, du2023preformer, cao2023tempo} for disentangling underlying patterns within deep features.

\subsubsection{Basis Expansion} 
Basis expansion is a mathematical method used to represent a function or a set of data points in terms of a new set of pre-defined functions. These new functions form the bases for a function space, allowing any function in that space to be expressed as a linear combination of these basis functions. In time series analysis, basis expansion is used to reveal nonlinear temporal relationships by decomposing the time series into a combination of basic variations, thereby enhancing interpretability. As a representative, N-BEATS~\cite{oreshkin2019nbeats} presents hierarchical decomposition to time series by utilizing a fully connected layer to produce expansion coefficients for both backward and forward forecasts. For the $l$-th block in the proposed hierarchical architecture, the operation is
\begin{equation}
    \begin{aligned}
        \mathbf{X}_l &= \mathbf{X}_{l-1} - \hat{\mathbf{X}}_{l-1} \\
         \hat{\mathbf{X}}_{l},  \hat{\mathbf{Y}}_{l} &= \operatorname{Block}_l (\mathbf{X}_l),
    \end{aligned}
\label{equ: nbeats}
\end{equation}
where $\hat{\mathbf{X}}_{l-1}$ is the backcast results which restrict the block to approximate the input signal $\mathbf{X}_{l-1}$, then $\mathbf{X}_{l}$ removes the portion of well-estimated signal $\hat{\mathbf{X}}_{l-1}$ from $\hat{\mathbf{X}}_{l-1}$, therefore providing a hierarchical decomposition. $\hat{\mathbf{Y}_{l}}$ is the partial forecast based on the decomposed input $\mathbf{X}_{l}$ and the final forecast $\hat{\mathbf{Y}} = \sum_l \hat{\mathbf{Y}}_l$ is the sum of all partial forecasts.

Subsequently, N-HiTs \cite{challu2023nhits} redefines N-BEATS by incorporating subsampling layers before the fully connected blocks. This modification enhances input decomposition through multi-frequency data sampling and improves future prediction via multi-scale interpolation. DEPTS~\cite{fan2022depts} proposes a novel decoupled formulation for periodic time series by introducing a periodic state as a hidden variable. It then develops a deep expansion module atop residual learning to perform layer-by-layer expansions between observed signals and these hidden periodic states. Similarly, DEWP \cite{fan2024dewp} is a stack-by-stack expansion model designed to handle multivariate time series data. Each stack within the model comprises a variable expansion block for capturing dependencies among multiple variables and a time expansion block for learning temporal dependencies.

\subsubsection{Matrix Factorization} 
The aforementioned two decomposition methods primarily address univariate series or are applied to multivariate series in a variate-independent way. Here, we explore factorization-based decomposition that is particularly pertinent to multivariate time series. These high-dimensional data can be formalized as a matrix where rows represent variates and columns denote time points. Since variables in multivariate time series tend to be highly correlated, they can be reduced to a more compact space. Matrix factorization methods \cite{xiong2010temporal} achieve this by decomposing the high-dimensional series data into the product of two matrices in a lower-dimensional latent space. 
For a multivariate time series $\mathbf{X} \in \mathbb{R}^{T\times C}$, the matrix can be approximated by the multiplications of two lower rank embedding matrix, $\mathbf{X} \approx \mathbf{F}\hat{\mathbf{X}}$, in which $\mathbf{F} \in \mathbb R ^ {T \times k}$, $\hat{\mathbf{X}} \in \mathbb R ^ {k \times C}$ and $k$ is a hyperparameter.

Besides the estimation, regularizers are critical in factorization-based models to mitigate overfitting. Going beyond the canonical design that takes the squared Frobenius norm as regularizers, temporal regularized matrix factorization (TRMF) \cite{yu2016temporal} designs an autoregressive-based temporal regularizer to describe temporal dependencies among latent temporal embeddings. Further, \cite{takeuchi2017autoregressive} extended TRMF with a new spatial autoregressive regularizer to estimate low-rank latent factors by simultaneously learning the spatial and temporal autocorrelations. For enhanced modeling of the trends and seasonality of real-world time series data, NoTMF \cite{chen2022nonstationary} integrates the vector autoregressive process with differencing operations into the classical low-rank matrix factorization framework. Addressing the challenge of parameter tuning, BTF \cite{chen2021bayesian} offers a fully Bayesian approach, uniting probabilistic matrix factorization and the vector autoregressive process within a single probabilistic graphical model. Diverging from autoregressive-based temporal regularization, DeepGLO \cite{sen2019think} leverages a temporal convolution network to capture non-linear dependencies. Similarly, LSTM-GL-ReMF \cite{yang2021real} employs an LSTM-based temporal regularizer for complex long-term and short-term non-linear temporal correlations, complemented by a graph Laplacian spatial regularizer \cite{cai2010graph} for capturing spatial correlations.

\subsection{Fourier Analysis}
Fourier analysis \cite{sneddon1995fourier, goodman2005introduction} is a powerful mathematical technique that transforms a physical signal into the frequency domain, thereby illuminating the inherent periodic properties of the original data. It has long been recognized as a fundamental analytical tool across diverse domains. Given that time series data are typically recorded as sequences of discrete observations sampled from continuous signals, Fourier analysis has naturally emerged as a mainstream approach in time series modeling with favorable effectiveness and efficiency \cite{bloomfield2004fourier, samiee2014epileptic}. Introducing the Fourier domain not only enhances the representation of the original series but also provides a global perspective on temporal dynamics. The resulting frequency spectrum distribution offers crucial insights into the essential periodic components and underlying cyclical behaviors of time series. In practice, algorithms such as Fast Fourier Transform (FFT) \cite{brigham1988fast} and Wavelet Transform (WT) \cite{meyer1992wavelets} act as foundational bridges, connecting the discrete temporal domain to the frequency domain. These techniques have gained increasing popularity in the modular design of deep time series models \cite{wang2018multilevel,yao2019stfnets,rhif2019wavelet,minhao2021t, yang2022unsupervised,wang2023wavelet,yang2023waveform, yi2023survey}. Existing approaches can be broadly categorized into two main types: time-domain and frequency-domain modeling.

\subsubsection{Time-Domain Modeling} 
The fundamental principle behind the Fourier transform is that sequential data can be decomposed and represented by a series of periodic signals. Consequently, it can be used to identify potentially dominant periods and their corresponding frequencies in the data by analyzing the highest amplitude components. As a prominent integration, TimesNet \cite{wu2023timesnet} employs the Fast Fourier Transform (FFT) to extract the most significant frequencies with the highest amplitude values. It then reshapes the 1D time series data into a 2D space based on these identified periods, thereby facilitating hierarchical representation learning. Building upon TimesNet, PDF \cite{dai2023periodicity} posits that frequencies with larger values facilitate a more discernible distinction between long-term and short-term relationships. 

The frequency-domain representation provides valuable information regarding both amplitudes and phases. Low-frequency components, for instance, correspond to slower variations or underlying trends in the signal, while high-frequency components capture finer details or rapid fluctuations. A significant body of work has focused on leveraging frequency-domain information to enhance the model's capability in capturing temporal dependencies. FiLM \cite{zhou2022film} combines Fourier analysis with low-rank approximation and introduces a frequency enhanced layer (FEL). This approach retains portions of the representation related to low-frequency Fourier components and the top eigenspace, thereby effectively reducing noise and accelerating training speed. FITS \cite{xu2023fits} integrates a low-pass filter (LPF) to eliminate high-frequency components above a specified cutoff frequency, thereby compressing the model size while preserving essential information. Taking an opposing view, FEDformer \cite{zhou2022fedformer} posits that retaining only low-frequency components is insufficient for time series modeling, as it risks dismissing important data fluctuations. Based on this consideration, FEDformer aims to capture a global view of time series by representing the series using a randomly selected constant number of Fourier components, encompassing both high-frequency and low-frequency contributions.

Beyond merely exploiting sequence information obtained from Fourier transforms, some research focuses on achieving efficient computation through the Fast Fourier Transform. Auto-correlation is a fundamental concept in time series analysis that measures the dependence between observations at different time points within a sequence of data. The Wiener-Khinchin theorem \cite{wiener1930generalized} provides a mathematical relationship between the auto-correlation function and the power spectral density (PSD) of a stationary random process, where the auto-correlation function represents the inverse Fourier transform of the PSD. Taking the data as a real discrete-time process, Autoformer \cite{wu2021autoformer} proposes an Auto-Correlation mechanism with an efficient Fast Fourier Transform to capture the series-wise correlation. 

\subsubsection{Frequency-Domain Modeling} 
Building on advancements in time-frequency analysis from signal processing, several approaches have been developed to analyze time series simultaneously in both the time and frequency domains. ATFN \cite{yang2020adaptive}, for instance, utilizes an augmented sequence-to-sequence model to learn trending features of non-stationary time series, alongside a frequency-domain block for capturing periodic patterns. TFAD \cite{zhang2022tfad} introduces a time-frequency analysis-based model that employs temporal convolutional networks to obtain both time-domain and frequency-domain representations.

Some works have developed specialized deep learning architecture to process the frequency domain of time series. STFNet \cite{yao2019stfnets} applies Short-Time Fourier Transform to input signals and applies filtering, convolution, and pooling operations directly in the frequency domain. StemGNN~\cite{cao2020spectral} combines Graph Fourier Transform (GFT) and Discrete Fourier Transform to model both inter-series correlations and temporal dependencies. EV-FGN \cite{yi2022edge} uses a 2D discrete Fourier transform on the spatial-temporal plane of the embeddings and performs graph convolutions for capturing the spatial-temporal dependencies simultaneously in the frequency domain. FreTS \cite{yi2024frequency} leverages Discrete Fourier Transform (DFT) to transform the data into the frequency domain spectrum and introduces frequency domain MLPs designed for complex numbers with separated modeling for the real parts and the imaginary parts. FCVAE \cite{wang2024revisiting} integrates both the global and local frequency features into the condition of Conditional Variational Autoencoder (CVAE) concurrently. Recent TSLANet \cite{eldele2024tslanet} proposes a lightweight Adaptive Spectral Block (ASB) to replace the self-attention mechanism, which is achieved via Fourier-based multiplications by global and local filters. FourierDiffusion \cite{crabbé2024time} extends the score-based SDE formulation of diffusion models to complex-valued data and therefore implements time series diffusion in the frequency domain.

\begin{figure*}[t]
    \centering
    \includegraphics[width=\linewidth]{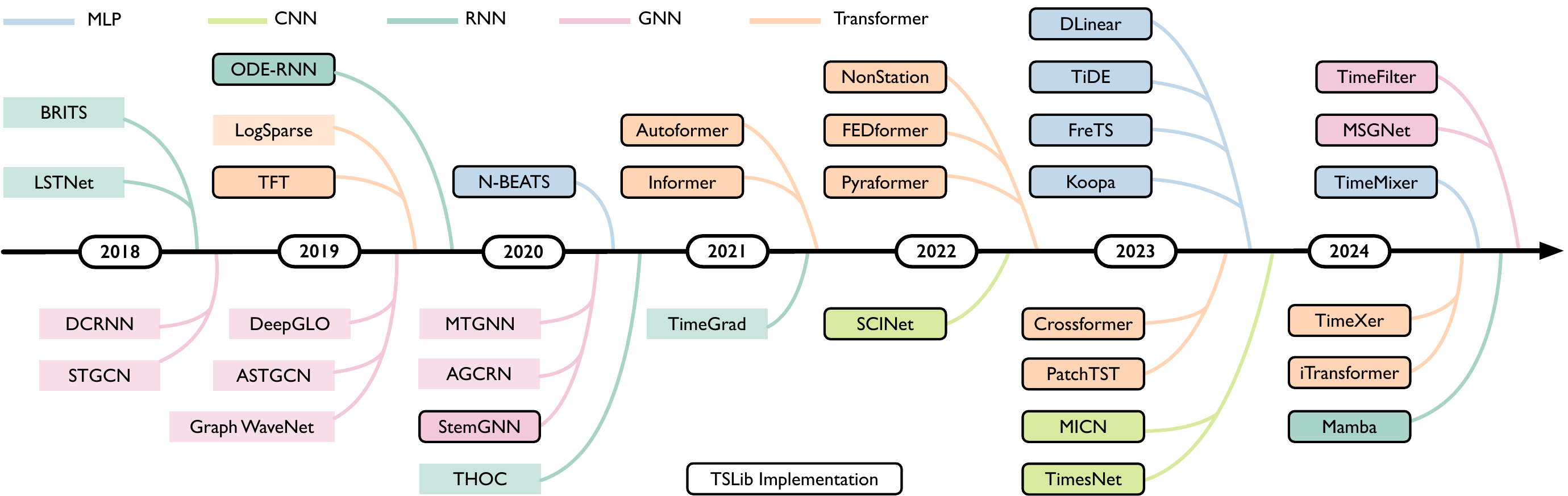}
    \vspace{-15pt}
    \caption{An overview of representative time series models in chronological order. We mark models with different colors based on their architectures.}
    \label{fig:overview}
    \vspace{-10pt}
\end{figure*}

\section{Model Architectures}\label{sec:arch}

As discussed in Section \ref{sec:basic}, time series models aim to unearth the intrinsic temporal dependencies and variate correlations to uncover underlying patterns and dynamics within observations. Existing approaches can be broadly categorized into five groups based on their architectural backbones (Figure~\ref{fig:overview}), namely, MLP-based, RNN-based, CNN-based, GNN-based, and Transformer-based. Each of these architectures possesses distinct strengths and design principles that address specific challenges in time series analysis. In this section, we provide a comprehensive technical review of current deep time series models, specifically highlighting their distinct approaches to time series modeling.

\subsection{Multilayer Perceptrons}
As a cornerstone of traditional statistical time series models, the Auto-regressive (AR) model posits that the model output depends linearly on its own past values. Inspired by the notable efficacy of auto-regressive models, Multilayer Perceptrons (MLP) have become a popular architecture for time series modeling.

As a prime example of MLP-based models, N-BEATS \cite{oreshkin2019nbeats} integrates no time-series-specific components to capture temporal patterns. Specifically, as described in Eq. \eqref{equ: nbeats}, N-BEATS consists of deep stacks of fully-connected layers with two residual branches in each layer, one for the backcast prediction and the other for the forecast branch. Expanding the idea of neural basis expansion, N-HiTs \cite{challu2023nhits} enhances prediction through multi-rate signal sampling and hierarchical interpolation, while N-BEATSx \cite{olivares2023neural} further extends the framework by incorporating exogenous variables.

\begin{figure}[h]
    \centering
\includegraphics[width=0.9\linewidth]{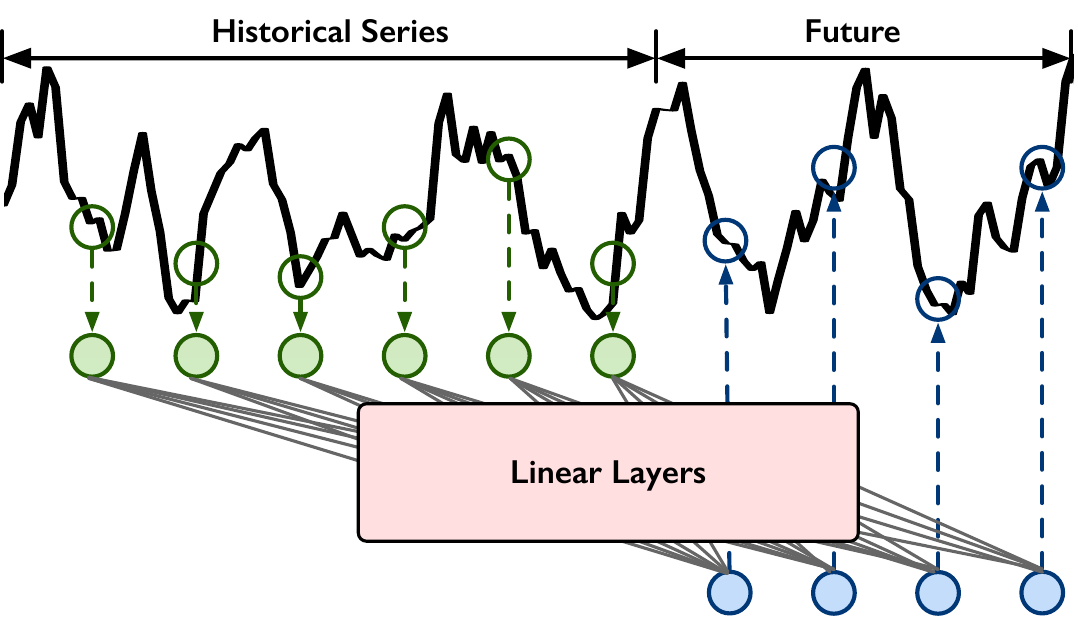}
    \vspace{0pt}
    \caption{Illustration of a basic MLP-based model in forecasting task, which captures the future-past dependencies with learnable MLP parameters.}
    \label{fig:MLP}
    \vspace{0pt}
\end{figure}

Recent research, notably DLinear \cite{zeng2023transformers}, rethinks the effectiveness of complicated deep architectures for temporal modeling. It argues a simple linear regression in the raw space that achieves remarkable performance in both modeling and efficiency. As illustrated in Figure~\ref{fig:MLP}, many prevalent MLP-based deep time series models, primarily designed for forecasting, rely on such simple linear layers. Also lightweight but effective, FITS \cite{xu2023fits} advocates that time series forecasting can be treated as interpolation exercises within the complex frequency domain. It introduces a complex-valued linear layer to learn amplitude scaling and phase shift directly in the frequency domain. Inspired by MLP-Mixer~\cite{tolstikhin2021mlp} in computer vision, several works have attempted to leverage MLPs to model both temporal and variate dependencies. TSMixer \cite{chen2023tsmixer} contains interleaving time-mixing and feature-mixing MLPs to extract information from different perspectives. To better capture global dependencies, FreTS \cite{yi2024frequency} investigates the learned patterns of frequency-domain MLPs, which operate on both inter-series and intra-series scales to capture channel-wise and time-wise dependencies in multivariate data.

Moving beyond simple linear layers over discrete time points, TimeMixer suggests that time series exhibit distinct patterns across different sampling scales and introduces an MLP-based multiscale mixing architecture. TiDE \cite{das2023long} incorporates exogenous variables to enhance time series prediction. Based on Koopman theory and Dynamic Mode Decomposition (DMD) \cite{schmid2010dynamic}, which is a dominant approach for analyzing complicated dynamical systems, Koopa \cite{liu2023koopa} hierarchically disentangles dynamics through an end-to-end predictive training framework and can utilize real-time incoming series for online development.

\subsection{Recurrent Neural Networks}
Recurrent Neural Networks (RNNs) are specifically designed for sequential data \cite{hochreiter1997long, sutskever2014sequence, cho2014learning}, such as natural language processing \cite{nallapati2016abstractive} and audio modeling \cite{mehri2016samplernn}.  Given the inherent sequential nature of time series, RNNs have naturally become a popular choice for time series analysis \cite{tealab2018time}. Existing RNN-based deep time series models primarily focus on combating the gradient vanishing problem caused by the vanilla recurrent structure and modeling the mutual correlation among multivariate variables. 

\begin{figure}[h]
    \centering
    \includegraphics[width=0.9\linewidth]{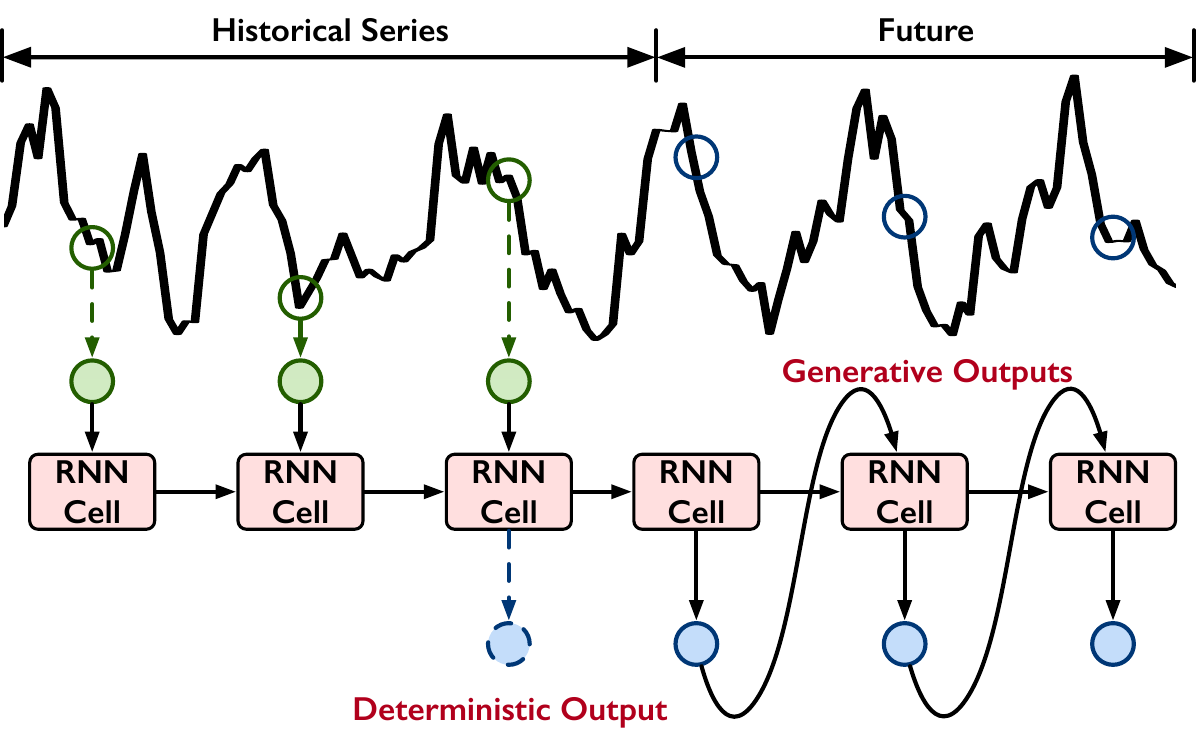}
    \vspace{-5pt}
    \caption{Illustration of RNN-based model in the forecasting task.}
    \label{fig:RNN}
    \vspace{0pt}
\end{figure}

Numerous prior works \cite{cao2018brits, yoon2018estimating,che2018recurrent, baytas2017patient, shen2020timeseries} have leveraged various RNN variants to capture complex dependencies.
LSTNet \cite{lai2018modeling} combines the recurrent structure with the convolutional layer to capture both the short-term local dependency between variables and long-term patterns for time series. It further introduces a novel recurrent-skip component, based on periodic patterns, to alleviate gradient vanishing in modeling long-term dependencies. Similarly, DA-RNN \cite{qin2017dual} combines the recurrent unit with a dual-stage attention mechanism to adaptively extract relevant series at each time step. Beyond deterministic forecasts, DeepAR \cite{salinas2020deepar} proposes an auto-regressive recurrent network to predict the probability distribution of further time points. Technologically, it learns not only the seasonal behavior with time series but dependencies on given covariates, allowing the model to make predictions even when there is little or no historical data. Recent work \cite{wang2023wavelet} introduces a novel DTWAtt module, which integrates a data-dependent wavelet function into the BiLSTM network as wavelet attention to analyze dynamic frequency components in nonstationary time series.

Based on Markovian state representation, the State Space Model (SSM) \cite{gu2021combining} is a classical mathematical framework for capturing probabilistic dependencies between observed measurements in stochastic dynamical systems.
While SSMs have proven their effectiveness and efficiency in processing well-structured time series data, traditional approaches have to refit each series separately and therefore cannot infer shared patterns from a dataset of similar time series. With the advent of deep learning, modern SSMs are often implemented in a recurrent manner. By adapting and propagating a deterministic hidden state, RNNs offer an alternative to classical state space models by representing long-term dependencies in continuous data. Consequently, some works \cite{fraccaro2016sequential,krishnan2017structured} have sought to fuse classical state space models with deep neural networks. Representative among these, the Deep State Space Model (DSSM) \cite{rangapuram2018deep} leverages a recurrent neural network (RNN) to parametrize a specific linear SSM, thereby incorporating structural assumptions while learning complex patterns. The Structured State Space sequence model (S4) \cite{gu2022efficiently} introduces a novel parametrization for the SSM by conditioning matrix $\mathbf{A}$ with a low-rank correction, which allows for stable diagonalization and significantly enhances the model's long-term modeling capacity. More recently, Mamba \cite{gu2023mamba} has emerged as a powerful method for modeling long sequential data with linear complexity w.r.t. sequence length. Utilizing a simple selection mechanism that parameterizes the SSM parameters based on input, Mamba can discern valuable information in observations, posing an effective way for time series analysis.

Going beyond learning among discretized time steps, Neural Ordinary Differential Equations (Neural ODEs) \cite{chen2018neural} extend RNNs into a continuous-time version. Technically, Neural ODEs parameterize the derivative of hidden states through neural networks, enabling seamless modeling of temporal evolution in a differentiable manner. Given the ability to model continuous-time, several works have explored applying Neural ODEs to handle irregularly-sampled time series data \cite{rubanova2019latent, kidger2020neural}. 
Building on these advancements, \cite{yin2021augmenting} integrates Neural ODEs with physical priors to augment physical models with deep networks for forecasting complex dynamics. These capabilities make Neural ODEs a compelling complement to RNN-based methods, offering a more flexible and robust approach for time series analysis.

\subsection{Convolutional Neural Networks}
Since the semantic information of time series is mainly hidden in the temporal variation, Convolutional neural networks (CNN) \cite{he2016deep, gu2018recent} have become a competitive backbone.
Their strength lies in their ability to capture local features and recognize intricate patterns, which has been demonstrated with remarkable success in various computer vision tasks, including image classification \cite{tan2019efficientnet}, segmentation \cite{minaee2021image}, and object detection \cite{redmon2016you}.

Considering the temporal continuity of time series data, previous works \cite{le2016data, ismail2020inceptiontime} apply a one-dimensional CNN (1D CNN) to discern the local patterns of time series data. Recent SCINet\cite{liu2022scinet} employs canonical convolutions with a hierarchical downsample-convolve-interact architecture to capture dynamic temporal dependencies across varying temporal resolutions. Inspired by the idea of masked convolution \cite{van2016pixel}, Wavenet\cite{oord2016wavenet} introduces causal convolution and dilated causal convolution to model long-range temporal causality. Similarly, Temporal Convolutional Networks (TCN) \cite{bai2018empirical} uses a stack of dilated convolutional kernels with progressively enlarged dilation factors to achieve a large receptive field. However, the limited receptive field of TCN often hinders its capacity to capture global relationships in time series data. Based on TCN, MICN\cite{wang2022micn} is a local-global convolution network that combines different convolution kernels to model temporal correlation from a local and global perspective. ModernTCN \cite{luo2024moderntcn} enhances the traditional TCN by separately employing Depthwise Separable Convolutions (DWConv) and Convolutional Feed-Forward Networks (ConvFFN) to capture cross-time and cross-variable dependencies, respectively. 

\begin{figure}[h]
    \centering
    \vspace{-5pt}
    \includegraphics[width=1\linewidth]{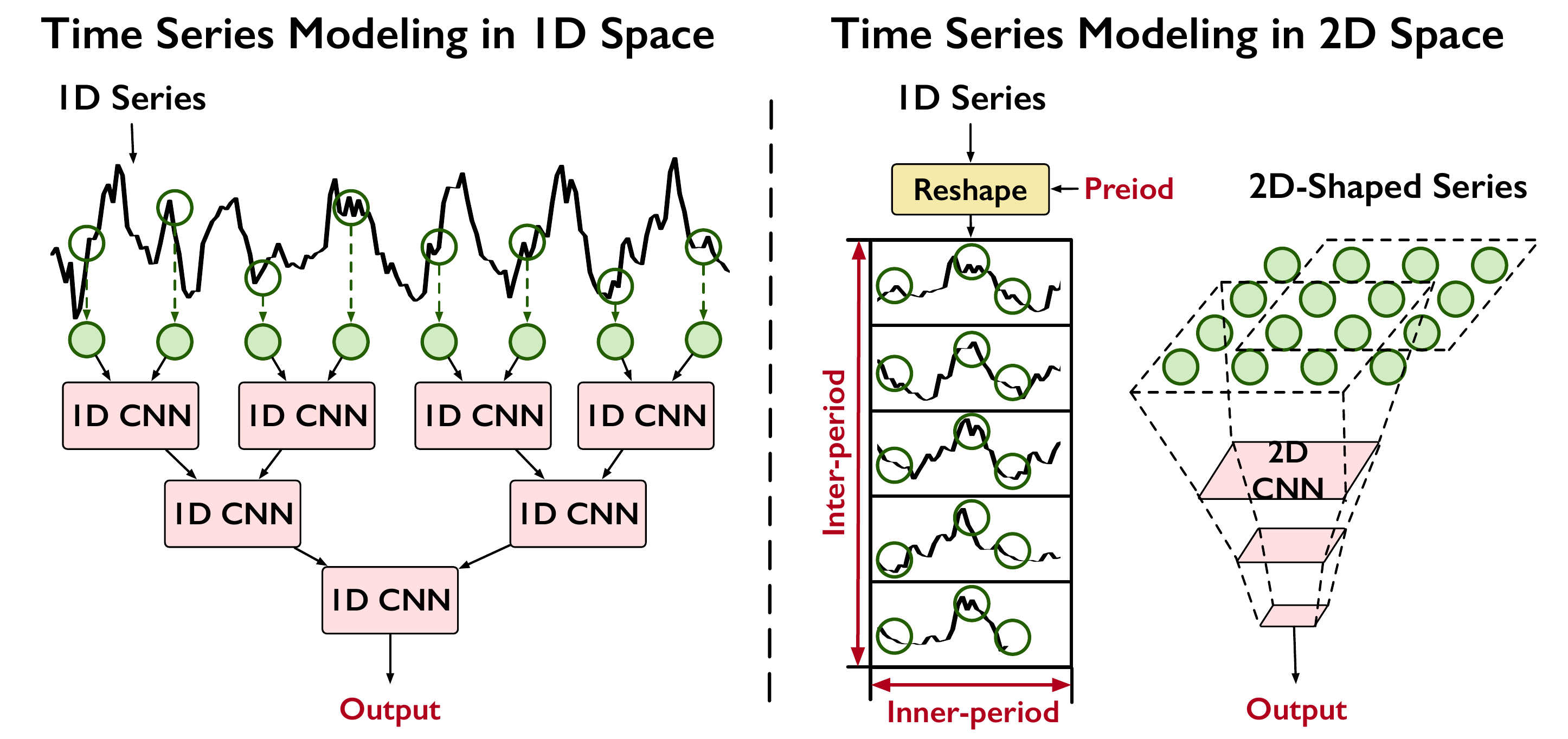}
    \vspace{-10pt}
    \caption{Comparison of existing CNN-based time series models from the perspective of representation space.}
    \vspace{0pt}
    \label{fig:CNN}
\end{figure}

Motivated by the inherent periodicity of time series data, TimesNet \cite{wu2023timesnet} goes beyond traditional 1D modeling. It \cite{wu2023timesnet} transforms the 1D time series $\mathbf{X_{\text{1D}}}$ data into a set of 2D tensors $\mathbf{X}_{\text{2D}} = \{\mathbf{X}^1_{\text{2D}}, \cdots, \mathbf{X}^k_{\text{2D}}\}$ in each TimesBlock based on the estimated period lengths, where the inter-period variations are presented in tensor columns and inner-period ones are shown in tensor rows. Here $k$ is a hyperparameter, corresponding to multiple 1D-to-2D transformations with different periods. Subsequently, TimesNet applies an Inception Block \cite{szegedy2015going, szegedy2016rethinking} to process these transformed 2D tensors, which can be summarized as:
\begin{equation}
  \begin{aligned}
\mathbf{X}^{i}_{\text{2D}} &=\operatorname{Reshape}\left(\operatorname{Padding}(\mathbf{X}_{\text{1D}})\right),\ i\in\{1,\cdots, k\}\\
\hat{\mathbf{X}}^{i}_{\text{2D}} &=\operatorname{Inception}\left(\mathbf{X}^{i}_{\text{2D}}\right),\ i\in\{1,\cdots, k\}\\
\hat{\mathbf{X}}^{i}_{\text{1D}} &=\operatorname{Trunc}\left(\operatorname{Reshape}\big(\hat{\mathbf{X}}^{i}_{\text{2D}}\big)\right),\ i\in\{1,\cdots, k\},\\
  \end{aligned}
\end{equation}
where $\mathbf{X}^i_{\text{2D}}$ is the $i$-th transformed 2D tensor. After passing through the inception block $\operatorname{Inception}(\cdot)$, the learned 2D representations are transformed back to 1D for aggregation. These transformations enable TimesNet to effectively capture both multi-scale intraperiod-variation and interperiod-variation simultaneously. Furthermore, by leveraging hierarchical convolutional layers, TimesNet is capable of learning both high-level and low-level representations, facilitating comprehensive time series analysis across four distinct tasks.

\subsection{Graph Neural Networks}
\begin{figure}[h]
    \vspace{-5pt}
    \centering
    \includegraphics[width=\linewidth]{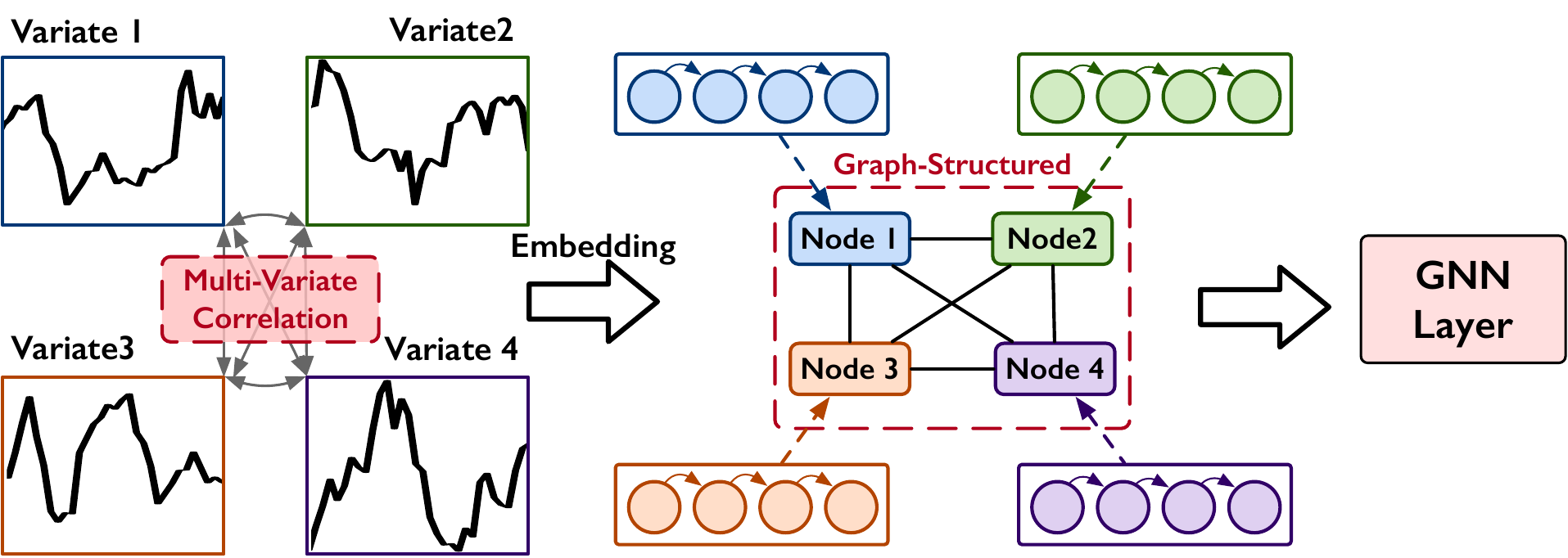}
    \vspace{-10pt}
    \caption{Illustration of modeling multivariate time series using GNN.}
    \vspace{-0pt}
    \label{fig:GNN}
\end{figure}

Analyzing multivariate time series data is often challenging due to the complicated and often nonlinear correlations between variables. To address this, Graph neural networks (GNNs) \cite{scarselli2008graph, kipf2017semi} have been widely adopted in time series analysis. As depicted in Figure \ref{fig:GNN}, by modeling multivariate data as a spatiotemporal graph, where each node represents a variable, GNNs can extract relationships among neighboring nodes and capture the temporal evolution of node attributes over time, providing a framework for understanding the underlying dynamics of multivariate time series.

The primary goal of GNN architecture is to model the latent topological relations in multivariate data. Accordingly, existing GNN-based works can be broadly categorized into two types: those that rely on a pre-defined graph structure as input and those that learn the graph structure dynamically. DCRNN \cite{li2018dcrnn_traffic} models the spatial dependency of traffic as a diffusion process on a directed graph and uses diffusion convolution to capture the spatial dependency, alongside a recurrent neural network to capture temporal dynamics. Similarly, STGCN \cite{yu2018spatio} integrates graph convolutional networks to model the spatial dependencies among traffic sensors with temporal convolutions to capture the temporal dependencies in the traffic time series data. Graph WaveNet \cite{ijcai2019p264} combines graph convolution with dilated causal convolution and learns an adaptive dependency matrix through node embedding, enabling the model to automatically capture hidden spatial dependencies in spatial-temporal graph data. Similarly, AGCRN \cite{bai2020adaptive} enhances the traditional graph convolutional network with node adaptive parameter learning and data-adaptive graph generation modules, allowing for the automatic capture of spatial and temporal correlations without a pre-defined graph structure. MTGNN \cite{wu2020connecting} introduces a graph learning layer to adaptively learn the graph adjacency matrix, thereby capturing hidden relationships among multivariate time series data. STFGNN \cite{li2021spatial} employs a Spatial-Temporal Fusion Graph Neural Network with a generated temporal graph to learn localized spatial-temporal heterogeneity and global spatial-temporal homogeneity. StemGNN \cite{cao2020spectral} leverages the advantages of both the Graph Fourier Transform (GFT) and the Discrete Fourier Transform (DFT), modeling multivariate time series in the spectral domain.

\subsection{Transformers}

\begin{figure}[h]
\vspace{-5pt}
    \centering
    \includegraphics[width=\linewidth]{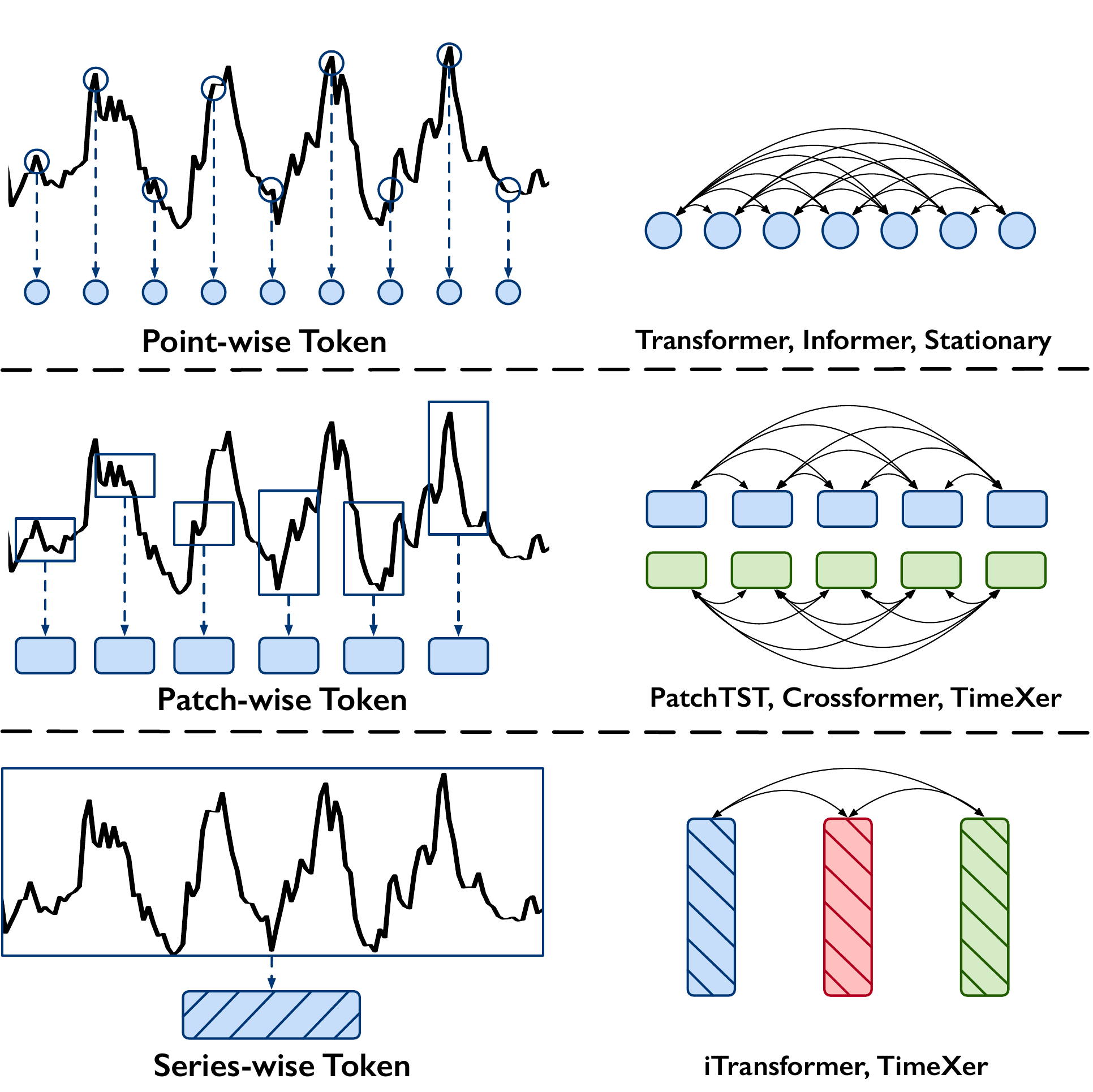}
    \vspace{-20pt}
    \caption{Comparison of different types of tokenization used for Transformer-based time series models and corresponding representative work, ranging from point-wise, patch-wise to series-wise.}
    \vspace{-5pt}
    \label{fig:transformer}
\end{figure}

In view of the great success in the field of natural language processing \cite{devlin2018bert,lu2019vilbert, yang2019xlnet, brown2020language, zhao2023survey} and computer vision \cite{dosovitskiy2020vit,touvron2021training,liu2021swin,wang2021pyramid}, Transformers have emerged as a powerful backbone for time series analysis. Benefiting from the self-attention mechanism \cite{vaswani2017attention}, Transformer-based models excel at capturing long-term temporal dependencies and entangled multivariate correlations. As overviewed in Figure \ref{fig:transformer}, existing Transformer-based time series models can be categorized based on the granularity of representation used in the attention mechanism, namely point-wise, patch-wise, and series-wise approaches. 

\begin{table*}[th]
  \setlength{\abovecaptionskip}{0cm}
  \setlength{\belowcaptionskip}{0cm}
  \vspace{-5pt}
\caption{Model cards of Transformer-based deep time series models with architectural details.}
  \label{tab:transformer_card}
  \centering
  \begin{small}
  \renewcommand{\multirowsetup}{\centering}
  \renewcommand\arraystretch{1.2}
    \begin{tabular}{llccl}
    \toprule
    Category & Method  & Architecture & Embeddding      & Attention Mechanism  \\ 
    \midrule
    Vanilla  & Transformer & Enc-Dec  & Standard & $\operatorname{Full Attention}(\mathbf{Q}, \mathbf{K}, \mathbf{V}) = \operatorname{Softmax}(\frac{\mathbf{QK}^\intercal}{\sqrt{d}})\mathbf{V}$ \\ \midrule
    \multirow{6}{*}{Point-wise} & \multirow{2}{*}{LogSparse \cite{li2019enhancing}} & \multirow{2}{*}{Dec-only} &  \multirow{2}{*}{Standard}   & $\mathbf{\widehat{Q}}, \mathbf{\widehat{K}} = \operatorname{CausalCov}(\mathbf{H})$    \\ 
    & &  & & $\operatorname{Full Attention}(\mathbf{\widehat{Q}}, \mathbf{\widehat{K}}, \mathbf{V}) = \operatorname{Softmax}(\frac{\mathbf{\widehat{Q}\widehat{K}}^\intercal}{\sqrt{d}})\mathbf{V}$   \\
     & \multirow{2}{*}{Informer \cite{zhou2021informer}} & \multirow{2}{*}{Enc-Dec} &  \multirow{2}{*}{Standard}   & $\overline{\mathbf{Q}} = \operatorname{Top}_u \left(\overline{M}(\mathbf{q}_i, \mathbf{K}) \right)$     \\ 
     & &  & & $\operatorname{ProbSparse-Attention}(\mathbf{Q}, \mathbf{K}, \mathbf{V}) = \operatorname{Softmax}(\frac{\mathbf{\overline{Q}K}^\intercal}{\sqrt{d}})\mathbf{V}$  \\ 
    &Pyraformer \cite{liu2021pyraformer} & Enc-Dec  &  Standard   & $\operatorname{Pyramid-Attention}(\mathbf{Q}, \mathbf{K}, \mathbf{V}) = \operatorname{Masked}(\operatorname{Softmax}(\frac{\mathbf{QK}^\intercal}{\sqrt{d}})\mathbf{V}) $         \\ \midrule
    \multirow{4}{*}{Patch-wise} &  Autoformer \cite{wu2021autoformer} & Enc-Dec & Standard & $\mathrm{Auto\text{-}Correlation}(\mathbf{Q},\mathbf{K},\mathbf{V})= \sum_{i=1}^{k}\mathrm{Roll}(\mathbf{V},\tau_{i})\widehat{\mathcal{R}}_{\mathbf{Q},\mathbf{K}}(\tau_{i})$    \\
    &  Crossformer \cite{zhang2023crossformer} & Enc-Dec & Patch-Wise & $\operatorname{Full Attention}(\mathbf{Q}, \mathbf{K}, \mathbf{V}) = \operatorname{Softmax}(\frac{\mathbf{QK}^\intercal}{\sqrt{d}})\mathbf{V}$    \\ 
     &  PatchTST \cite{Yuqietal-2023-PatchTST} & Enc-only & Patch-Wise &  $\operatorname{Full Attention}(\mathbf{Q}, \mathbf{K}, \mathbf{V}) = \operatorname{Softmax}(\frac{\mathbf{QK}^\intercal}{\sqrt{d}})\mathbf{V}$   \\ \midrule
    Variate-wise &  iTransformer \cite{liu2023itransformer} & Enc-only & Variate-Wise & $\operatorname{Full Attention}(\mathbf{Q}, \mathbf{K}, \mathbf{V}) = \operatorname{Softmax}(\frac{\mathbf{QK}^\intercal}{\sqrt{d}})\mathbf{V}$                 \\ \bottomrule
  \end{tabular}
  \end{small}
  \vspace{-10pt}
\end{table*}

\subsubsection{Point-wise Dependency}
Due to the serial nature of time series, most existing Transformer-based works learn point-wise representations and apply attention mechanisms to capture the correlations among different time points. Among these point-wise modeling approaches, Data Embedding stands as a crucial component, mapping raw time series values into a high-dimensional representation. Given time series $\mathbf{X} \in \mathbb R ^{T \times C}$ with corresponding time stamp information $\mathbf{X}^{\rm{ts}} \in \mathbb R ^{T \times D}$, where $C$ is the variate number and $D$ is the types of time stamps, the embedding module can be summarized as:
\begin{equation}
    \begin{aligned}
        \mathbf{H}_t &= \operatorname{Projection}(\mathbf{X}_t) + \operatorname{PE}(\mathbf{X}_t) + \operatorname{TE}(\mathbf{X}^{\rm{ts}}_t)),
    \end{aligned}
\end{equation}
where $\mathbf{H}_t\in\mathbb{R}^{T\times d_{\rm{model}}}$ and $d_{\rm{model}}$ is the dimension of the emebedded representation, value projection $\operatorname{Projection}: \mathbb{R}^C \mapsto \mathbb{R}^{d_{\rm{model}}}$ and timestamp embedding $\operatorname{TE}: \mathbb{R}^D \mapsto \mathbb{R}^{d_{\rm{model}}}$ are implemented by channel-dimension linear layers, and $\operatorname{PE(\cdot)}$ denotes the absolute position embedding to preserve the sequential context of input series.

To optimize the application of Transformer architecture to the time series domain, existing literature primarily falls into two categories: designing pre-processing modules and modifying the attention mechanisms. As previously discussed in Section 3.1, techniques like RevIN \cite{kim2021reversible} and Stationary \cite{liu2022non} have achieved superior performance by integrating Normalization and De-Normalization modules before and after the Transformer blocks. Notably, Stationary \cite{liu2022non} further introduces De-stationary Attention to mitigate the problem of over-stationarization.

Given that canonical attention leads to a quadratic computational complexity, various efficient attentions~\cite{zhou2021informer,wu2021autoformer,liu2021pyraformer, zhou2022fedformer} have been proposed to mitigate the complexity arising from point-wise modeling, which is summarized in Table~\ref{tab:transformer_card}.
LogSparse \cite{li2019enhancing} proposes \textit{Convolutional Self-Attention} to replace canonical attention by employing causal convolutions to produce queries and keys in the self-attention layer. Informer \cite{zhou2021informer} introduces a \textit{Query Sparsity Measurement}, where a larger value indicates a higher chance of containing the dominant information in self-attention. Building on this, it designs ProbSparse self-attention, which selectively uses only the top queries with the highest measurement results, thereby significantly reducing quadratic complexity in both time and memory. Pyraformer \cite{liu2021pyraformer} constructs a multi-resolution tree and develops a \textit{Pyramidal Attention Mechanism}, in which every node can only attend to its neighboring, adjacent, and children nodes.  With a calculated mask for attention, Pyraformer captures both short- and long-temporal dependencies with linear computational complexity.

\subsubsection{Patch-wise Dependency}
Patch-based architectures play a crucial role in the Transformer models for both Natural Language Processing (NLP) \cite{devlin2018bert} and Computer Vision (CV) \cite{dosovitskiy2020vit}. Since point-wise representations often fall short in capturing local semantic information in temporal data, several studies \cite{Yuqietal-2023-PatchTST, zhou2024one, liu2024timer} have been devoted to exploring patch-level temporal dependencies in time series data.

Pioneer work Autoformer \cite{wu2021autoformer} proposes an \emph{Auto-Correlation} \emph{Mechanism} to replace canonical point-wise self-attention, enabling the model to learn series-wise dependencies. Based on the stochastic process theory \cite{papoulis2002probability}, Auto-Correlation utilizes the Fast Fourier Transform to discover the time-delay similarities between different sub-series. A time delay module is further proposed to aggregate similar sub-series from underlying periods instead of the relation between scattered points, which firstly explores the sub-series level modeling in Transformer-based models.

Different from modifying the attention mechanism, most recent works utilize patch-wise representations of the data and perform a canonical self-attention mechanism to capture patch-wise dependencies \cite{Yuqietal-2023-PatchTST, du2023preformer, xue2024card}. PatchTST \cite{Yuqietal-2023-PatchTST} and its subsequent works split time series $\mathbf{X}$ into a sequence of overlapped patches and embed each patch following:
\begin{equation}
    \begin{aligned}
        \{\mathbf{P}_1, \mathbf{P}_2, \cdots, \mathbf{P}_N\} &= \operatorname{Patchify} \left(\mathbf{X}\right), \\
        \mathbf{H}_{i} &= \operatorname{PatchEmbed}(\mathbf{P}_i) + \mathbf{E}_{\rm pos}^{i}. \\
    \end{aligned}
\end{equation}
Let $P$ denote the patch length and $N$ the corresponding number of patches, and $\mathbf{P}_i$ is the $i$-th patch of length $P$. The patches are mapped to the latent space through a temporal linear projection $\operatorname{PatchEmbed}: \mathbb{R}^ P \mapsto \mathbb{R}^{d_{\rm{model}}}$ and a learnable position embedding $\mathbf{E}_{\rm pos} \in \mathbb{R} ^ {d_{\rm{model}} \times N}$. Based on the vanilla attention mechanism, PatchTST \cite{Yuqietal-2023-PatchTST} learns the patch-wise dependencies. Going beyond PatchTST, Pathformer \cite{chen2023multi} introduces a multi-scale Transformer that employs adaptive pathways to dynamically select patch sizes, thereby learning multi-scale temporal representations.

The success of PatchTST also benefits from its channel-independence design, where each temporal patch-level token encapsulates information from only a single series. In addition to capturing the patch-level temporal dependencies within one single series, recent approaches \cite{du2023preformer, xue2024card} have endeavored to capture interdependencies among patches from different variables over time. Crossformer \cite{zhang2023crossformer} introduces a \textit{Two-Stage Attention layer} containing a Cross-Time Stage and a Cross-Dimension Stage to efficiently capture the cross-time and cross-variate dependencies between each patch token. For the obtained embedded vector $\mathbf{H} \in \mathbb R ^ {N \times C \times d_{\rm model}}$, the overall attention stage can be described as follow:
\begin{equation}
    \begin{aligned}
        \mathbf{Z}^{\rm time} &= \operatorname{Attention}^{\rm time} \left( \mathbf{H}, \mathbf{H}, \mathbf{H} \right) \\
        \mathbf{B} &= \operatorname{Attention}_1^{\rm dim} \left (\mathbf{R}, \mathbf{Z}^{\rm time}, \mathbf{Z}^{\rm time} \right) \\
        \mathbf{\overline{Z}}^{\rm dim} &= \operatorname{Attention}_2^{\rm dim} \left (\mathbf{Z}^{\rm time}, \mathbf{B}, \mathbf{B} \right), \\
    \end{aligned}
\end{equation}
where $\mathbf{R} \in \mathbb R ^{\mathbf{N} \times \mathbf{C} \times d_\text{model}}$ is a learnable vector array used as a router to gather information from all dimensions and then distribute the gathered information.

\subsubsection{Series-wise Dependency}
Further expanding the receptive field, recent works tokenize the entire time series to capture inter-series dependencies.  iTransformer~\cite{liu2023itransformer} introduces $\operatorname{VariateEmbed}$ for multivariate data and for $i$-th variable $\mathbf{X}^{(i)}$, it can be formulated as:
\begin{equation}
 \begin{aligned}
     \mathbf{H}^{(i)} &= \operatorname{VariateEmbed}(\mathbf{X}^{(i)})\\
\end{aligned}
\end{equation}
where $\operatorname{VariateEmbed}: \mathbb{R}^T \to \mathbb{R}^{d_{\rm model}}$ is instantiated as trainable linear projector.
Based on the global representations of each series, iTransformer utilizes the vanilla Transformer without any architectural modifications to capture mutual correlations in multivariate time series data. Similarly, TimeXer\cite{wang2024timexer} focuses on forecasting with exogenous variables and utilizes patch-level and series-level representations for endogenous and exogenous variables, respectively. Additionally, an endogenous global token is introduced to TimeXer, which serves as a bridge in-between and therefore captures intra-endogenous temporal dependencies and exogenous-to-endogenous correlations jointly.

\section{Time Series Foundation Models}\label{sec:ltsm}

With the advent of Large Language Models (LLMs), the large-scale pre-training paradigm has inspired our community moving towards Time Series Foundation Models (TSFMs) to learn complex temporal dependencies and capture the underlying dynamics inherent in large-scale time series data. ForecastPFN \cite{dooley2023forecastpfn} introduces the first zero-shot forecasting model trained exclusively on a novel synthetic data distribution. Recently, numerous large-scale time series models have been proposed, primarily leveraging Transformer-based architectures. These models are pre-trained on massive time series datasets, often at the billion- or trillion-scale, to achieve strong zero-shot generalization for downstream tasks. Technologically, these models can be broadly categorized into two types: time series native foundation models built from scratch and adapted from LLMs.

\subsection{Time-Series Native TSFM}

\textbf{Encoder-based TSFM}
{Inspired by the success of encoder-only deep time series forecasters, some encoder-based TSFMs have been proposed. MOMENT \cite{goswami2024moment} is a Transformer encoder pre-trained with a mask reconstruction objective on univariate time series data, which can be fine-tuned for various downstream analysis tasks. Moirai \cite{woo2024unified} introduces a multivariate pre-training paradigm by flattening variates and considering all variates as a single sequence. The successor Moirai-MoE \cite{liu2024moirai} utilizes sparse experts to model diverse time series patterns in a data-driven manner, enabling automatic token-level specialization. Additionally, researchers have explored TSFMs within alternative architectural frameworks. TTMs \cite{ekambaram2024tiny} employs a lightweight architecture consisting of Multi-Layer Perceptrons and integrates techniques such as adaptive patching, diverse resolution sampling, and resolution prefix tuning to facilitate pre-training on datasets with varied resolutions while maintaining minimal model capacity. TiRex \cite{auer2025tirex}, based on xLSTM, incorporates a novel masking strategy that retains state-tracking capabilities, allowing pre-trained time series models to provide reliable uncertainty estimates over extended prediction horizons.

\begin{figure}[t]
    \centering\includegraphics[width=\linewidth]{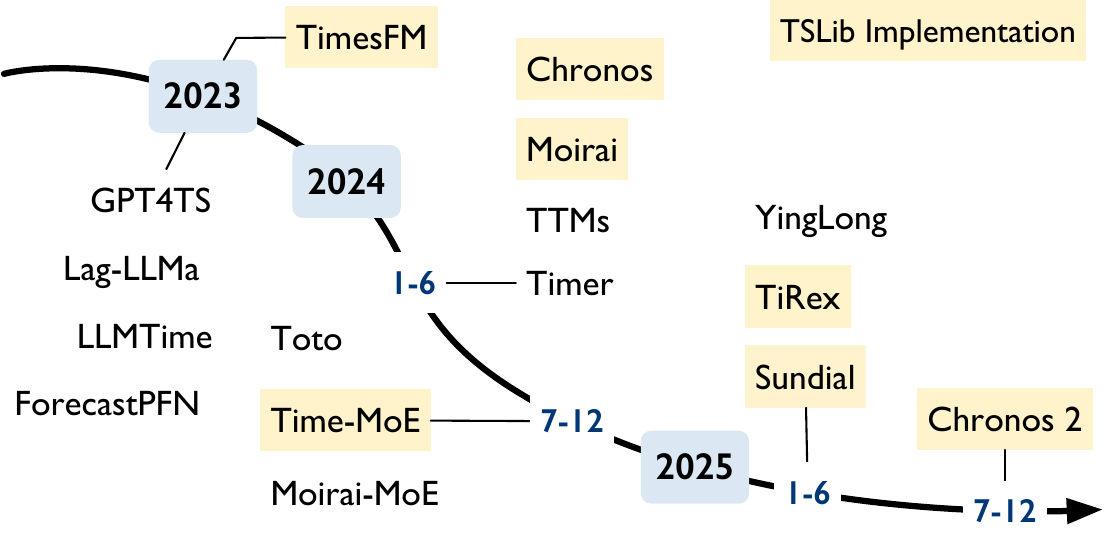}
    \vspace{-15pt}
    \caption{Recent development of time series foundation models. The timeline is established according to the release date of the paper or technical report for the model.}
    \vspace{-10pt}
    \label{fig:ltsm-timeline}
\end{figure}

\textbf{Decoder-based TSFM}
Motivated by the substantial progress of decoder-only LLMs, another line of research adopts decoder-only architectures for rollout generation. Lag-Llama \cite{rasul2023lag} is a decoder-only Transformer that uses time series lags as covariates. TimesFM \cite{das2024decoder} is a decoder-style attention model with input patching, using a large time-series corpus comprising both real-world and synthetic datasets. Going beyond forecasting tasks, Timer \cite{liu2024timer} converts forecasting, imputation, and anomaly detection of time series into a unified generative task and proposes a generative pre-trained Transformer for general time series analysis. Time-MoE \cite{shi2024time} introduces a scalable, unified architecture with a sparse mixture-of-experts design, extending the training corpus to 300 billion time points. Previous works rely on the inherent generative ability of the pre-trained Transformer architecture. Recently, Sundial \cite{liu2025sundial} proposes TimeFlow based on flow-matching generation to predict next-patch’s distribution, allowing Transformers to be trained without discrete tokenization and make probable predictions. Pretrained on one trillion time points, Sundial demonstrates superior performance on both point and probabilistic forecasting with a fast inference speed. Tailored for high-dimensional observability, Toto \cite{cohen2024toto} introduces an advanced attention mechanism that allows for efficient grouping of multivariate time series features, reducing computational overhead while maintaining high accuracy.

\subsection{LLM Empowered TSFM}
Based on the similar sequential nature, fine-tuning pre-trained language models to equip them with time series analysis capabilities has become a promising research topic. When applying LLMs to time series, it is essential to tokenize the time series before feeding it to a pre-trained model.

GPT4TS \cite{zhou2024one} proposes a unified framework for diverse time series analysis tasks by using a pre-trained GPT-2 model and fine-tuning the positional embeddings and the parameters of the layer normalization for each analysis task. LLMTime \cite{gruver2023llmtime} introduces a tokenization scheme that encodes real-valued data as a string of digits after fixing the numerical precision and scaling the data appropriately. Based on the T5 family, Chronos \cite{ansari2024chronos} introduces a pre-trained probabilistic time series forecaster on a large collection of publicly available datasets, complemented by a synthetic dataset. Technologically, Chronos tokenizes time series values using scaling and quantization into a fixed vocabulary and trains the model on these tokenized time series via the cross-entropy loss. Recently, Chronos2 \cite{ansari2025chronos} extends univariate forecasting to universal forecasting by incorporating a novel group attention mechanism.

\subsection{Discussion}
The selection between time series native foundation models and LLM empowered TSFMs hinges on the trade-off between specialized temporal modeling and generalized reasoning. Native TSFMs are fundamentally designed to capture fine-grained temporal patterns, offering superior fidelity to intricate variations via architectures tailored for numerical data. Consequently, they serve as the preferred paradigm for purely numerical scenarios, providing stable and efficient modeling of intrinsic dynamics to deliver rigorous point estimations. LLM empowered TSFMs, conversely, excel in cross-modal synthesis by leveraging inherent semantic priors and reasoning abilities of LLMs. These models are particularly advantageous for complex scenarios where contextual metadata, such as textual descriptions or event logs, can be aligned with temporal patterns or where internal world knowledge is required to facilitate modeling. Therefore, while native models define the frontier of dedicated numerical prediction, LLM empowered paradigms offer a versatile alternative for context-aware and cross-domain applications.

\section{Time Series Library}\label{sec:tslib}

\begin{figure*}[htb]
    \centering
    \includegraphics[width=\linewidth]{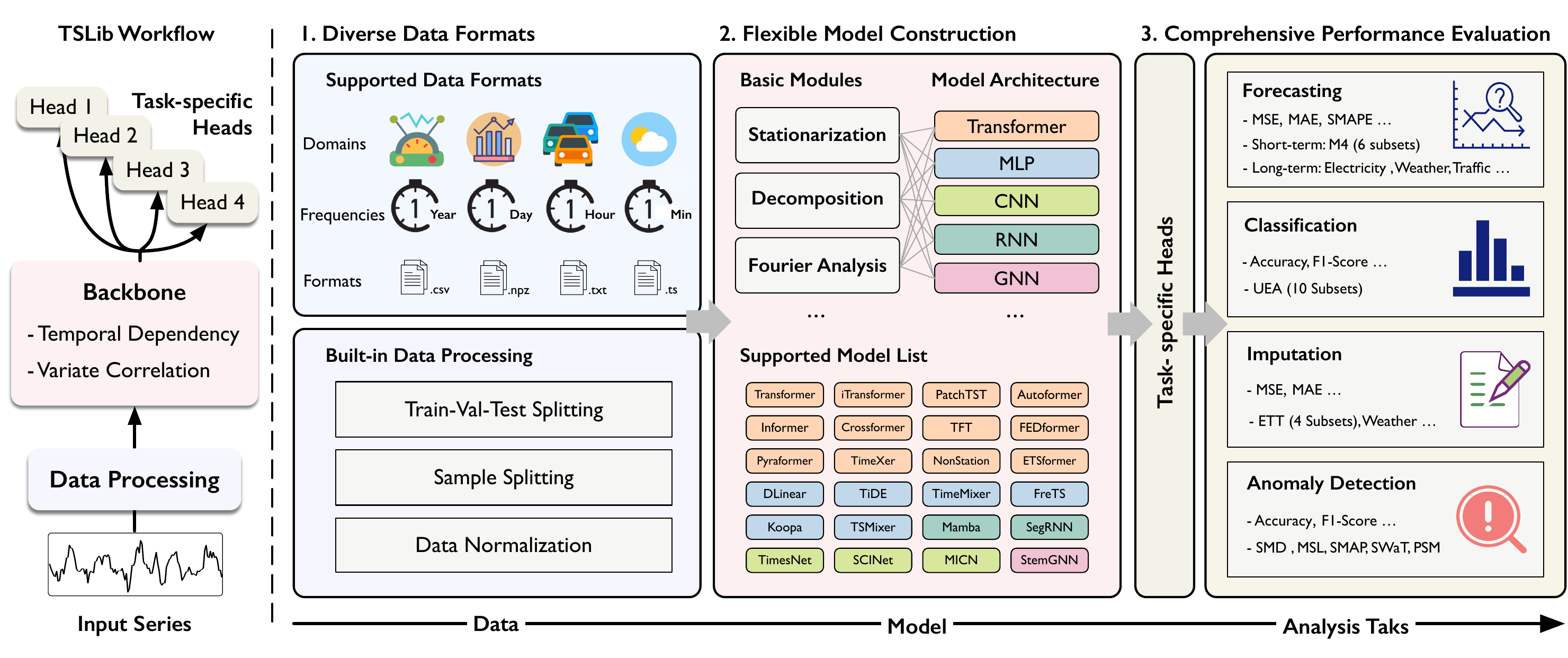}
    \vspace{-20pt}
    \caption{Overall framework of Time Series Library (TSLib). Left: Workflow for task-general experiments. Right: Key features and design principles.}
    \label{fig:pipeline}
    \vspace{-10pt}
\end{figure*}

Recently, extensive exploration of deep time series methods has resulted in significant advances. Given such booming development, the issue of fair benchmarking poses a pressing challenge for future research. 
Thus, focusing on time series analysis, several benchmarks have been proposed, such as DGCRN \cite{li2023dynamic}, LibCity \cite{wang2021libcity}, DL-Traff \cite{jiang2021dl}, TS-bench \cite{hao2021tsbench}, and BasicTS \cite{shao2023exploring}. Besides, Autoformer \cite{wu2021autoformer} proposed a standard long-term forecasting benchmark covering different practical applications. Further, to verify the generality of different time series analysis models, TimesNet~\cite{wu2023timesnet} builds a more comprehensive model generalization benchmark covering five mainstream time series analysis tasks. However, these benchmarks typically have some limitations. One issue with current time series benchmarks is their limited coverage of time series analysis tasks and specific domains, which limits their practical applications. Moreover, these benchmarks often fail to provide detailed discussions and comprehensive summaries of task types, model architectures, and specific baseline methods. As a result, they do not effectively guide the design of more efficient time series methods or drive further development in the field.

To address these challenges effectively, we introduce Time Series Library (TSLib), a benchmark designed for the fair and comprehensive evaluation of deep time series models across a wide range of analysis tasks. As shown in Figure \ref{fig:pipeline}, TSLib provides a unified experimental framework that includes standardized evaluation protocols, a rich and diverse collection of real-world datasets, state-of-the-art and widely-used time series models, and a streamlined process for experimental validation and analysis. Although numerous time series libraries have been proposed in recent years, TSLib stands out by including advanced models and supporting a wide range of time-series analysis tasks, which have earned widespread recognition from the community as shown in Table \ref{tab:lib_compare}. Specifically, we meticulously followed the official codes and implemented 41 widely used and advanced deep time series models. Users can choose from these models based on their specific practical usage scenarios. The code is available at \href{https://github.com/thuml/Time-Series-Library}{https://github.com/thuml/Time-Series-Library}.


\begin{table}[t]
  \caption{Comparison between TSLib and other related libraries. $^\ast$ denotes the absence of deep models. ``Stars'' refers to GitHub stars as of Feb.~2026.}\label{tab:lib_compare}
  \setlength{\abovecaptionskip}{0cm}
  \setlength{\belowcaptionskip}{0cm}
  \vspace{-5pt}
  \centering
  \begin{small}
  \setlength{\tabcolsep}{0.8pt}
  \begin{tabular}{lccc}
    \toprule
     \scalebox{0.9}{Library} &  \scalebox{0.88}{\#Models}  &  \scalebox{0.9}{Task} & \scalebox{0.9}{Stars}\\
    \toprule
     \scalebox{0.9}{tslearn$^\ast$~\cite{tslearn_jmlr}} &  \scalebox{0.9}{10} &  \scalebox{0.8}{Clustering, Classification, Regression} & \scalebox{0.9}{3.1k} \\
     \scalebox{0.9}{StatsForecast$^\ast$~\cite{garza2022statsforecast}} &  \scalebox{0.9}{31} &  \scalebox{0.8}{Forecasting, Anomaly Detection} & \scalebox{0.9}{4.7k}\\
     \scalebox{0.88}{NeuralForecast~\cite{olivares2022library_neuralforecast}} &  \scalebox{0.9}{33}  &  \scalebox{0.8}{Forecasting} & \scalebox{0.9}{4.0k} \\
     \scalebox{0.9}{GluonTS~\cite{gluonts_jmlr}} &  \scalebox{0.9}{25} &  \scalebox{0.8}{Forecasting, Anomaly Detection} & \scalebox{0.9}{5.1k} \\
    \midrule
    \multirow{2}{*}{\scalebox{0.9}{TSLib (This Survey)}} & \multirow{2}{*}{ \scalebox{0.9}{41}}  &  \scalebox{0.8}{Forecasting, Imputation} & \multirow{2}{*}{\scalebox{0.9}{\scalebox{0.9}{11.6k}}} \\
    && \scalebox{0.8}{Classification, Anomaly Detection} \\
    \bottomrule
    \end{tabular}
    \end{small}
  \vspace{-15pt}
\end{table}

\subsection{Design and Implementation Principles} \label{sec:design}
TSLib is designed based on the well-established factory pattern and implements a unified interface between data, model, and experiment objects, thus enabling a clear separation between deep model creation and usage, promoting modularity and flexibility. By loading different data and model objects and combining task-specific heads, TSLib enables different datasets and models to be shared and extended, allowing easy switching between various datasets, models, and analysis tasks. Here are the key features. 

\textbf{Diverse Data Formats} TSLib natively supports a wide range of datasets in a variety of formats, including ``.csv'', ``.npz'', ``.txt'', etc. As summarized in Figure \ref{fig:pipeline}, TSLib currently supports more than 30 datasets with different sampled frequencies across four mainstream time series analysis tasks, all derived from real-world scenarios, such as energy, transportation, economics, weather, and medicine etc. Moreover, TSLib excels in scalability, allowing for the effortless integration of new data sources of different types. Besides, to support subsequent training and evaluation, TSLib includes a multitude of data processing steps, including time window splitting, data batch generation, etc, where the raw data is partitioned into separate windows for training, validation, and testing purposes, enabling streamlined model training and equitable comparisons. These steps serve as indispensable prerequisites for attaining precise and dependable results across a range of diverse time series analysis tasks.

Moreover, TSLib provides additional support for numerous crucial and effective data processing strategies \cite{passalis2019deep,kim2021reversible,liu2022non,cleveland1990stl} to enhance model performance and training efficiency. We encapsulate these various design strategies as basic data processing layers, encompassing techniques such as data normalization, time-frequency decomposition, Fourier analysis, and more. When utilizing TSLib, users have the flexibility to select these strategies to improve the training effect based on their specific requirements and objectives.

\textbf{Flexible Model Construction} 
As stated in the previous sections, we abstract the construction of deep time series models as the combination of basic modules (Section~\ref{sec:basic}) and model architectures (Section~\ref{sec:arch}). Specifically, we separate the most commonly used modules, such as stationarization and decomposition, from previous elaborately designed models. In this way, it is very convenient to reproduce or add a deep model with pre-defined basic modules. Further, although different models are originally proposed for different tasks, such as Autoformer \cite{wu2021autoformer} is for forecasting and TimesNet~\cite{wu2023timesnet} is for more generic tasks, following the ``model-centric'' principle in this survey, we unify diverse analysis tasks as a representation learning task with domain-specific heads, enabling a task-general evaluation for different backbones.

\textbf{Comprehensive Performance Evaluation}
Building upon the above two essential features, TSLib can naturally support the evaluation of diverse models on different tasks based on extensive datasets, building a comprehensive benchmark. Specifically, in TSLib, we provide evaluation for four mainstream time series analysis tasks: classification, imputation, forecasting, and anomaly detection. Each task involves specific evaluation metrics, enabling a comprehensive assessment of the performance of models. Noticing that, we further distinguish the forecasting task as long- and short-term tasks following the previous convention, where the separation depends on the forecasting horizon.

\subsection{Evaluation Protocols and Metrics} \label{sec:protocols}
To conduct a fair and comprehensive model comparison, TSLib provides standardized evaluation protocols for four mainstream time series analysis tasks following \cite{wu2023timesnet}. Specifically, different models are trained and tested based on the same training strategies and test sets to avoid potential unaligned configurations and ensure a rigorous comparison among different models.

Following previous research, for long-term forecasting and imputations, we adopt Mean Square Error (MSE) and Mean Absolute Error (MAE) as the primary evaluation metrics. For short-term forecasting, we use the Symmetric Mean Absolute Percentage Error (SMAPE) and Mean Absolute Scaled Error (MASE) as metrics, which focus on absolute errors and reduce the impact of outliers, providing reliable evaluations of forecast accuracy across different datasets and methodologies. In time series classification, we utilize classification accuracy as the evaluation metric. For anomaly detection, we employ the F1 score to validate the identification of abnormal values. The F1 score represents a balanced combination of precision and recall, offering a comprehensive assessment of a classifier's performance, especially when dealing with imbalanced abnormal classes.

\subsection{Supported Datasets}  \label{sec:datasets}

TSLib natively supports a variety of widely used datasets across different numbers of samples and variables, covering diverse domains and tasks. The correspondences between tasks and datasets are depicted in Figure \ref{fig:pipeline}. Since all datasets are commonly used, here is only a brief introduction, and more details can be found in previous papers \cite{wu2023timesnet}.

\textbf{Forecasting}
Forecasting is one of the most widely explored tasks in previous research, bringing us a wide choice of experimental datasets. Just as stated in design principles, we further divide this task into two types (long- and short-term) according to the forecasting horizon. For long-term time series forecasting, TSLib includes nine datasets, including Electricity Transformer Temperature (ETT, including four subsets) \cite{zhou2021informer}, Electricity, Weather, Traffic, Exchange, and Illness \cite{wu2021autoformer}. For the short-term forecasting task, we include the M4 dataset \cite{makridakis2020m4}, which comprises six sub-datasets with varying sampling frequencies and domains.

\textbf{Imputation }
Due to glitches, the collected time series data may contain partially missing values, posing a challenge for time series analysis. However, in real-world applications, it is really hard to obtain the ground truth values for the missing records. Thus, we utilize the Electricity Transformer Temperature (ETT, with four subsets) \cite{zhou2021informer}, Electricity, and Weather to construct the missing value situation by randomly masking a certain portion of time steps.

\textbf{Anomaly Detection }
Anomaly detection involves identifying unusual or abnormal patterns in a time series. These anomalies can indicate critical events, faults, or outliers that require attention or further investigation.
There are some mainstream anomaly detection datasets supported in TSLib, such as Server Machine Dataset (SMD) \cite{su2019robust}, Mars Science Laboratory rover (MSL) \cite{hundman2018detecting}, Soil Moisture Active Passive satellite (SMAP) \cite{hundman2018detecting}, Secure Water Treatment (SWaT) \cite{mathur2016swat}, and Pooled Server Metrics (PSM) \cite{DBLP:conf/kdd/AbdulaalLL21} which are collected from a variety of industrial scenarios. 

\textbf{Classification } We selected ten multivariate datasets from the UEA Time Series Classification Archive~\cite{Bagnall2018TheUM} to test the series-level classification tasks. These datasets cover a range of practical tasks, including gesture, action, audio recognition, and medical diagnosis through heartbeat monitoring. We pre-processed the datasets according to the descriptions in \cite{Zerveas2021ATF}.

\begin{figure*}[t]
    \centering
    \vspace{-5pt}
    \includegraphics[width=0.98\linewidth]{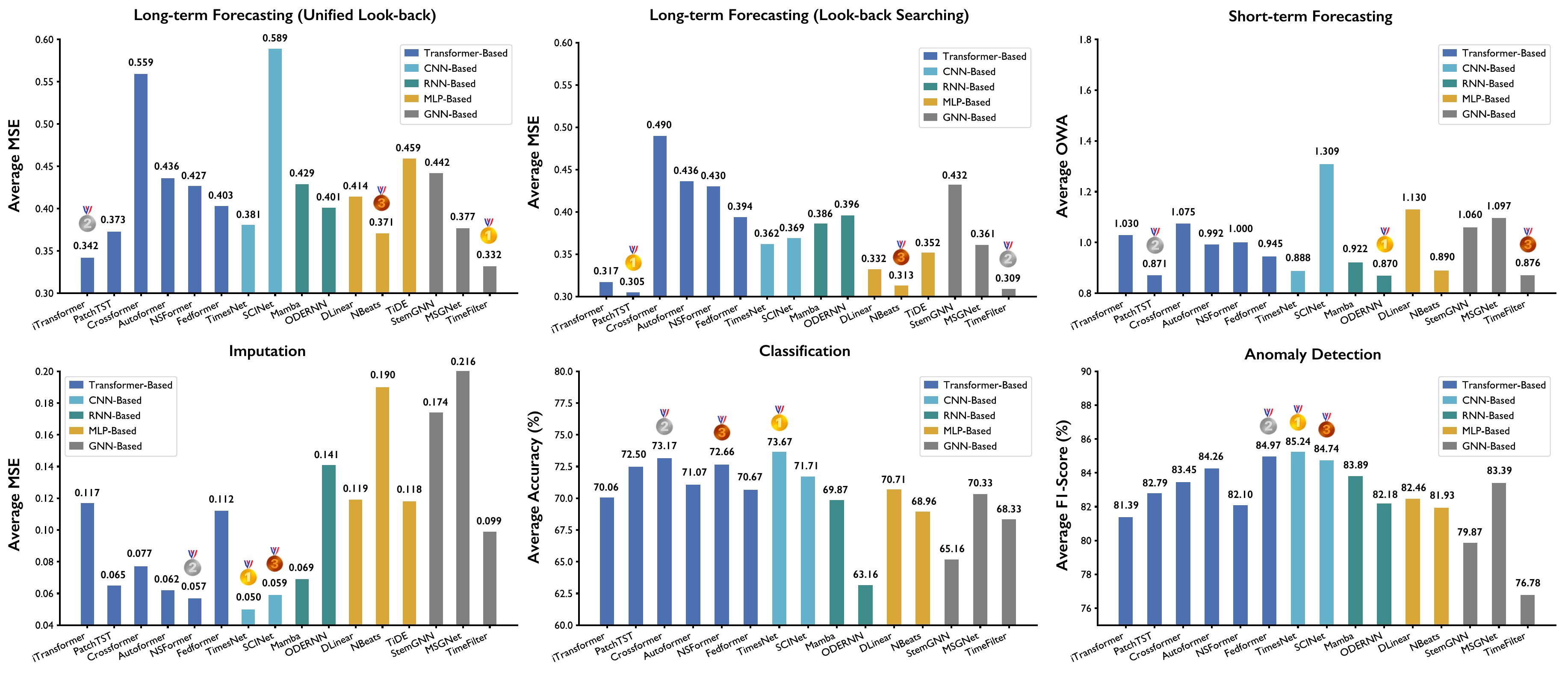}
    \vspace{-5pt}
    \caption{Comparison of model performance based on the Time Series Library. Full results are averaged from a diverse set of datasets supported by TSLib across four mainstream analysis tasks. In the classification and anomaly detection tasks, higher Accuracy or F1-score indicates better model performance. Conversely, lower average MSE or MAE demonstrates superior model effectiveness for the other tasks. The top three models for each time analysis task are highlighted on a leaderboard, and the optimal model for each time series analysis task is indicated.}
    \label{fig:main_result}
    \vspace{-10pt}
\end{figure*}

\subsection{Default Experimental Setups}\label{sec:exp_setting}
Here, we provide experimental setups for each analysis task supported by TSLib. For long-term forecasting and zero-shot forecasting, we use a look-back length of 36 and prediction lengths of $\{24, 36, 48, 60\}$ for the ILI dataset. For other datasets, a unified look-back length of 96 is adopted, with forecast horizons of $\{96, 192, 336, 720\}$. In short-term forecasting, the prediction lengths are set as $\{6, 8, 18, 13, 14, 48\}$ for yearly, quarterly, monthly, weekly, daily, and hourly data, respectively. In the classification task, we aim to categorize high-dimensional time series data into multiple classes.
In the imputation task, we randomly mask the time points in ratios of $\{12.5\%, 25\%, 37.5\%, 50\%\}$.
In the anomaly detection task, the reconstruction is a classical task for unsupervised point-wise representation learning, where the reconstruction error is a natural anomaly criterion. Therefore, we set deep time series models as the base models for reconstruction and utilize the reconstruction error as a shared anomaly criterion across all experiments.

\begin{table}[t]
  \caption{The default training configuration for each time series analysis task in Time Series Library. $\text{LR}^\ast$ denotes the initial learning rate.}\label{tab:model_config}
  \setlength{\abovecaptionskip}{0cm}
  \setlength{\belowcaptionskip}{0cm}
  \vspace{-5pt}
  \centering
  \begin{small}
  \setlength{\tabcolsep}{2.0pt}
  \begin{tabular}{lcccc}
    \toprule
    Task & LR$^\ast$ & Loss & BatchSize & Epochs\\
    \toprule
    Long-term Forecasting & $10^{-4}$ & MSE & 32 & 10 \\
    Short-term Forecasting & $10^{-3}$ & SMAPE & 16 & 10 \\
    Imputation & $10^{-3}$ & MSE & 16 & 10 \\
    Classification & $10^{-3}$ & Cross Entropy & 16 & 100 \\
    Anomaly Detection & $10^{-4}$ & MSE & 128 & 10 \\
    \bottomrule
    \end{tabular}
    \end{small}
  \vspace{-10pt}
\end{table}

\textbf{Implementation Details}
In this paper, all experiments were implemented using PyTorch and executed on an NVIDIA A100 GPU. The Adam optimizer was used for model optimization. Each model's performance was comprehensively evaluated across various time series analysis tasks. To achieve optimal performance and ensure fair comparisons across models and datasets, we conducted hyperparameter searches for each model architecture following \cite{wang2023timemixer}. Specifically, we search the number of hidden layers $L \in \{2, 3, 4\}$ and the dimension of hidden representations $d_{\text{model}}$ from the range $[2^4, 2^9]$. For each time series analysis task, TSLib utilized task-specific and model-shared training configurations, including learning rate, loss function, batch size, and number of epochs, which are listed in Table \ref{tab:model_config}.

\textbf{Look-back Searching Setting} Note that in long-term forecasting tasks, the model's performance is highly dependent on the historical length of the input series. Given the inconsistent look-back length of various papers, we further conducted an additional set of experiments referred to as ``Look-back Searching'' to ensure a fair and reliable comparison. In these experiments, we systematically searched for input length among $\{96, 192, 336, 512\}$ following \cite{wang2023timemixer} and performed an extensive hyperparameter search on each model's architecture to reach their optimal performance. This approach allowed us to assess the maximum achievable performance of each model while ensuring that the reported results accurately reflect their capabilities and facilitate a fair comparison across different models.

\begin{figure*}[t]
    \centering
    \includegraphics[width=\linewidth]{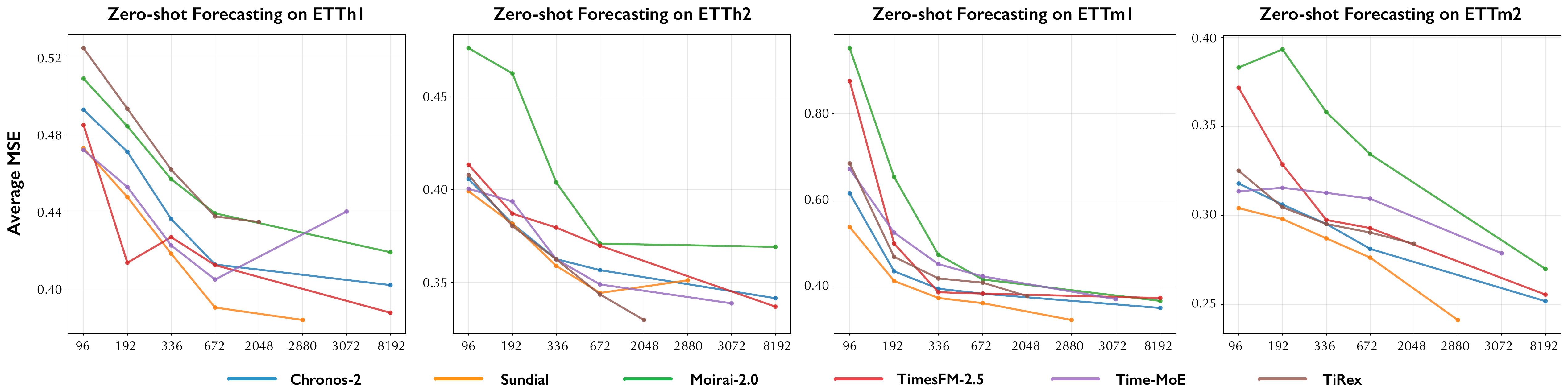}
    \vspace{-20pt}
    \caption{Zero-shot forecasting performance of different TSFMs with varying input lengths. For Moirai and TimesFM, whose maximum supported context exceeds the available dataset size, we align their input lengths with the maximum capacity of Chronos-2 to ensure a consistent and fair comparison.}
    \vspace{-10pt}
    \label{fig:zero-shot}
\end{figure*}

\subsection{Results and Analyses} \label{sec:main results}
To examine the strengths and limitations of various methods in mainstream time series analysis tasks, we select 16 representative deep models and 6 time series foundation models from TSLib for comparison. These models encompass five popular deep model architectures and tasks, including long-term and short-term forecasting, classification, imputation, and anomaly detection.

\textbf{Baselines} 
To thoroughly explore the effectiveness of different model architectures in various tasks, we conduct comparative experiments using state-of-the-art models designed based on different deep architectures. As shown in Figure \ref{fig:main_result}, we select several advanced and representative Transformer-based models: iTransformer \cite{liu2023itransformer}, PatchTST \cite{Yuqietal-2023-PatchTST}, Autoformer \cite{wu2021autoformer}, Non-Stationary Transformer (Stationary) \cite{liu2022non}, and FEDformer \cite{zhou2022fedformer} to compare the performance. Additionally, we consider TimesNet \cite{wu2023timesnet} and SCINet \cite{liu2022scinet} as the CNN-based models. For RNN-based models, we included the efficient Mamba \cite{gu2023mamba} and representative Neural ODE model, ODERNN \cite{rubanova2019latent}. The GNN-based baselines consist of  StemGNN \cite{cao2020spectral}, MSGNet \cite{cai2024msgnet}, and TimeFilter \cite{hu2025timefilter}, all of which enable a data-driven construction of dependency graphs. DLinear \cite{zeng2023transformers}, N-BEATS \cite{oreshkin2019nbeats}, and TiDE \cite{das2023long} are tested as representative MLP-based models. Since TiDE \cite{das2023long} depends on timestamps and cannot be easily adapted to tasks without timestamps in their datasets, such as short-term forecasting, anomaly detection, and classification, we only test it on the long-term forecasting task. We also include 6 mainstream time series foundation models, Moirai-2.0 \cite{liu2024moirai}, TimesFM-2.5 \cite{das2024decoder}, Time-MoE \cite{shi2024time}, TiRex \cite{auer2025tirex}, Sundial \cite{liu2025sundial}, and Chronos-2 \cite{ansari2025chronos} to evaluate their zero-shot forecasting performance. Since the ECL, Weather, and Traffic datasets were leaked during pre-training of TimesFM and other TSFMs, the evaluation was limited to the ETT datasets.

\textbf{Quantitative Results}
As illustrated in Figure \ref{fig:main_result}, Transformer-based models, notably iTransformer \cite{liu2023itransformer} and PatchTST \cite{Yuqietal-2023-PatchTST}, achieve state-of-the-art performance across diverse tasks, particularly in long-term forecasting, which underscores the critical role of specialized embedding and attention mechanisms in unlocking their architectural capacity. GNN-based models also demonstrate competitive efficacy, with TimeFilter \cite{hu2025timefilter} and MSGNet \cite{cai2024msgnet} excelling in forecasting and anomaly detection respectively, while TimesNet \cite{wu2023timesnet} stands out as a versatile milestone for task-general modeling. From an architectural perspective, the results reveal distinct trade-offs: MLP-based models offer surprising efficiency in forecasting but struggle with high-level representation learning, whereas CNN-based architectures exhibit more comprehensive capabilities across non-forecasting tasks. Although GNNs effectively capture inter-series correlations, they generally lack the task-general superiority of leading CNN frameworks. Meanwhile, RNN-based models remain viable for short-term tasks but show limited effectiveness in long-term predictions. Overall, the consistent superiority of Transformers across the benchmark highlights their formidable modeling capacity and potential as a unified backbone for advanced time series analysis.

Under the TSLib protocol, we also include a comparison of the zero-shot forecasting performance between the time series foundation models under varying input lengths. Specifically, we increase the input length from the conventional 96 to the maximum supported context of each model. Figure \ref{fig:zero-shot} reports the averaged results across four prediction horizons $\{96, 192, 336, 720\}$ on the ETT datasets. Overall, increasing the input context length generally yields improved zero-shot performance, underscoring the effectiveness of TSFMs in handling long contextual information. Notably, we observe that the performance of Sundial with an input length of 2880 surpasses that of other TSFMs at 8192. This suggests that excessively long inputs may introduce irrelevant information or stochastic noise that obscures meaningful temporal patterns. These findings indicate that optimal zero-shot performance is jointly determined by the model's architectural capacity and the inherent predictability of the input data, underscoring the importance of selecting an appropriate look-back window to balance context richness with data quality.

\begin{figure*}[t]
    \centering
    \includegraphics[width=\linewidth]{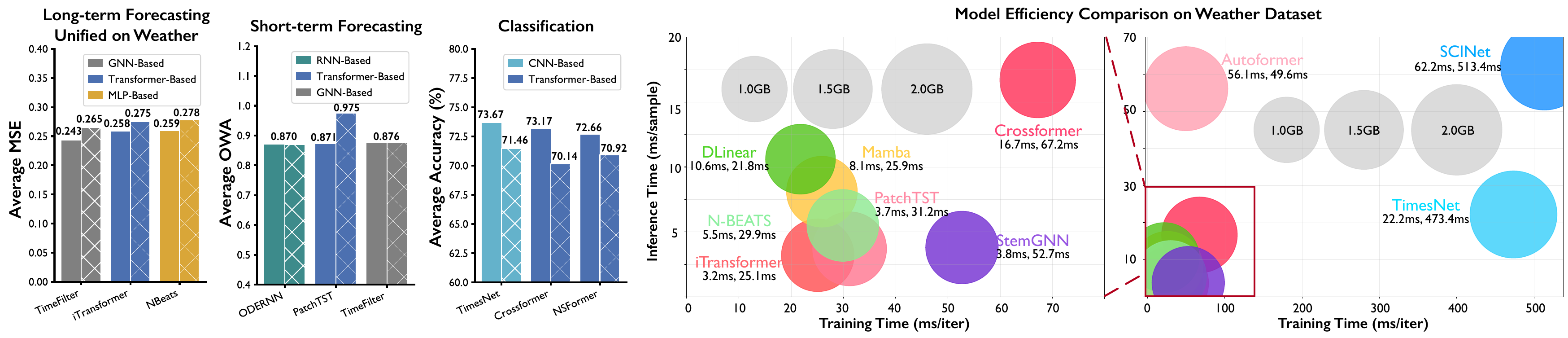}
    \vspace{-20pt}
    \caption{(Left) Robustness analysis on deep time series models, the cross-hatched bar indicates the performance under noisy data. (Right) Model efficiency tested for long-term forecasting on the Weather dataset.}
    \label{fig:qualitative-analysis}
    \vspace{-10pt}
\end{figure*}

\textbf{Qualitative Analysis}
We also provide an efficiency comparison to offer deeper insights into the practicality and feasibility of different approaches. Specifically, we measure the training time, inference time, and memory usage of each model in performing long-term forecasting, where all models operate under unified hyperparameter settings. As shown in Figure \ref{fig:qualitative-analysis} (Right), it is observed that the tokenization design affects the model efficiency, where Autoformer \cite{wu2021autoformer} with point-wise tokens is slower than PatchTST \cite{Yuqietal-2023-PatchTST} with patch-wise tokens and iTransformer \cite{liu2023itransformer} with series-wise tokens. Besides, CNN-based models, such as SCINet \cite{liu2022scinet} and TimesNet \cite{wu2023timesnet}, are much more time-consuming compared to other architectures since they have to adopt diverse kernels to capture intricate temporal patterns. In contrast, MLP-based models and the latest Mamba model stand out for their superior efficiency. Their simpler and more lightweight architectures make them particularly suitable for scenarios with limited computational resources. The above findings underscore the inherent trade-offs between model efficiency and predictive performance, which could also be a valuable guide for selecting models for resource-constrained scenarios.

To further assess the practical applicability of the surveyed models, we conducted a robustness analysis on the deep models. Given the inherent complexity and noise in real-world time series data, we systematically added random white noise to the input data to rigorously evaluate model robustness. Specifically, we focused on three representative analysis tasks: long-term forecasting, short-term forecasting, and classification, selecting the top three performing deep models for each task. As shown in Figure \ref{fig:qualitative-analysis} (Left),the introduction of noise leads to only marginal performance degradation across most models. This relative stability suggests that these models possess inherent robustness, confirming their viability and reliability in practical scenarios.

\section{Future Directions}\label{sec:future}
In this section, we present a discussion on promising directions for time series analysis. 

\vspace{-5pt}
\subsection{Time Series Pre-training }

The pretraining-finetuning paradigm, widely adopted in language \cite{qiu2020pre, devlin2018bert} and vision \cite{he2020momentum, liu2021self}, establishes foundational abilities through unsupervised pre-training before task-specific fine-tuning. Given the scarcity of labels, self-supervised pre-training \cite{liu2021self} has emerged as a powerful strategy for time series, primarily categorized into contrastive learning \cite{jaiswal2020survey} and masked modeling \cite{devlin2018bert, he2022masked}.

Contrastive learning regularizes representations by contrasting similar and dissimilar pairs \cite{jaiswal2020survey, chen2020simple}. To address the limitations of CV-based methods like SimCLR \cite{chen2020simple} in temporal modeling, specific approaches have been proposed: CPC \cite{oord2018cpc} captures temporal structure via predictive coding, TS2Vec \cite{Yue2022-TS2Vec} employs hierarchical instance- and patch-wise losses, and TS-TCC \cite{ijcai2021-324} introduces time series-specific augmentations. For long-term dependencies, CoST \cite{woo2022cost} learns disentangled seasonal-trend representations. Masked modeling is a reconstruction-based approach that predicts masked tokens from surrounding context \cite{baevski2022data2vec, devlin2018bert, he2022masked}. Following the BERT paradigm \cite{devlin2018bert}, TST \cite{zerveas2021transformer} first applied this to multivariate time series. Subsequent refinements focus on hierarchical and structural improvements: PatchTST \cite{Yuqietal-2023-PatchTST} operates at the patch level, while HiMTM \cite{zhao2024himtm} utilizes a hierarchical framework for multi-scale characteristics. Additionally, SimMTM \cite{dong2023simmtm} integrates manifold learning via neighborhood aggregation, and TimeSiam \cite{dong2024timesiam} adopts a Siamese network with asymmetric masking to capture intrinsic temporal correlations.

\subsection{Practical Applications}

\subsubsection{Probabilistic Forecasting}
Probabilistic forecasting is a pivotal advancement in practical applications of deep time series models, particularly for informed decision-making. Unlike previous widely-studied deterministic forecasting, probabilistic forecasting offers a distribution over potential future outcomes, enabling stakeholders to assess risks, evaluate confidence intervals, and make robust decisions under uncertainty.
Most existing probabilistic forecasting models rely on predefined probability distributions and are designed to predict the parameters of the distributions in the future. While these methods have achieved significant success, the dependence on predefined distributions can limit their ability to represent complex and heterogeneous real-world uncertainties. Therefore, accurately modeling and representing such complexities to generate more flexible and robust probabilistic forecasts remains a critical and active area of future research.

\subsubsection{Handling Extremely Long Series}
Deep time series models have demonstrated remarkable performance across a wide range of downstream tasks, but their applicability to longer time series data is often limited by scalability and high computational complexity. In industrial time series analysis, high-frequency sampling results in lengthy historical data, impeding the practical implementation of advanced deep models. Existing methods usually include patching techniques to enable them to handle long sequences, and when the input length becomes longer, the patch length can be increased accordingly to reduce the computational complexity. However, model performance is closely tied to patch length; hence, solely increasing patch size to reduce complexity may compromise capabilities. Therefore, addressing the limitations of deep models in handling longer time series could be a promising topic.

\subsubsection{Utilizing Exogenous Variables}
Since variations within the time series are often influenced by external factors, it is crucial to include exogenous variables in the analysis to gain a more comprehensive understanding of these factors. Exogenous variables, which are widely discussed in time series prediction tasks, are included in the model for uniform training in modeling time series data, without requiring separate analysis. However, in practical applications, different from multivariate time series analysis, the main variables and covariates usually occupy different positions. Given the crucial role played by exogenous variables in real-world applications, it is essential to explore a unified framework for modeling the relationships between the endogenous and exogenous variables, which allows for a more comprehensive understanding of interrelations and causality among different variants, leading to better and more reliable model performance, as well as interpretability.

\subsubsection{Processing Heterogeneous Data}
Modeling heterogeneous time series remains a significant challenge. This type of data exhibits diverse characteristics, including varying sampling rates, irregularities, and different length scales, which complicate the effective capture of underlying patterns. Furthermore, the requirement for fixed-size inputs in current deep learning models inherently limits their capability to handle this dynamic, variable-length nature. Addressing these issues demands innovative approaches that can both adapt to individual time series' unique features and capture overarching patterns across multiple series. This includes developing new feature extraction techniques and exploring alternative architectures suitable for variable-length inputs. Further research in this relatively unexplored area holds great promise for significant advances in fields like finance and healthcare, leading to deeper insights and enhanced predictive analytics capabilities.

\subsubsection{Handling Complex Disturbed Data}
Model robustness is essential for the reliable deployment of deep time series models in real-world applications. In practice, data is often subject to complex disturbances that go beyond typical stochastic fluctuations, including impulse spikes, non-stationary noise, and irregular missing values. Such disturbances can result from sensor instabilities, environmental changes, or intermittent transmission failures, creating highly dynamic and unpredictable data streams. The presence of these intricate disturbances often introduces transient distortions that can obscure underlying temporal patterns. Consequently, addressing these challenging data conditions requires the development of specialized modules designed to handle complex disturbances, making this a critical and promising direction for research. Progress in this area will not only improve the fundamental stability of time series analysis but also enable more dependable decision-making in safety-critical industrial systems.

\section{Conclusion}\label{sec:conclusion}
In this survey, we provide a systematic review of deep models in time series analysis and introduce Time-Series Library (TSLib) as a fair benchmark for deep time series models across various tasks. Compared with previous reviews that focus on a specific analysis task or model architecture, this paper provides a comprehensive survey and overview of existing deep models for time series analysis, ranging from forecasting, classification, imputation, and anomaly detection. By separating basic modules from complicated and diverse deep time series models, we successfully uncover the fundamental design principles for current research, forming an inspiring model-centric perspective. Besides, based on TSLib, which enables flexible and comprehensive model evaluation, we conduct extensive experiments to bring insights for model design and practical usage. With clear summarization, rich experiments, and in-depth discussion, we believe that this survey and TSLib can be instructive and useful for future research on deep time series models.

\section{Acknowledgments}

This work was supported by the National Natural Science Foundation of China (U2342217), the Fundamental and Interdisciplinary Disciplines Breakthrough Plan of the Ministry of Education of China (JYB2025XDXM803), and the National Engineering Research Center for Big Data Software.

\IEEEdisplaynontitleabstractindextext

\IEEEpeerreviewmaketitle


\bibliography{main}
\bibliographystyle{IEEEtran}

\vspace{-30pt}
\begin{IEEEbiography}[{\includegraphics[width=1in,clip,keepaspectratio]{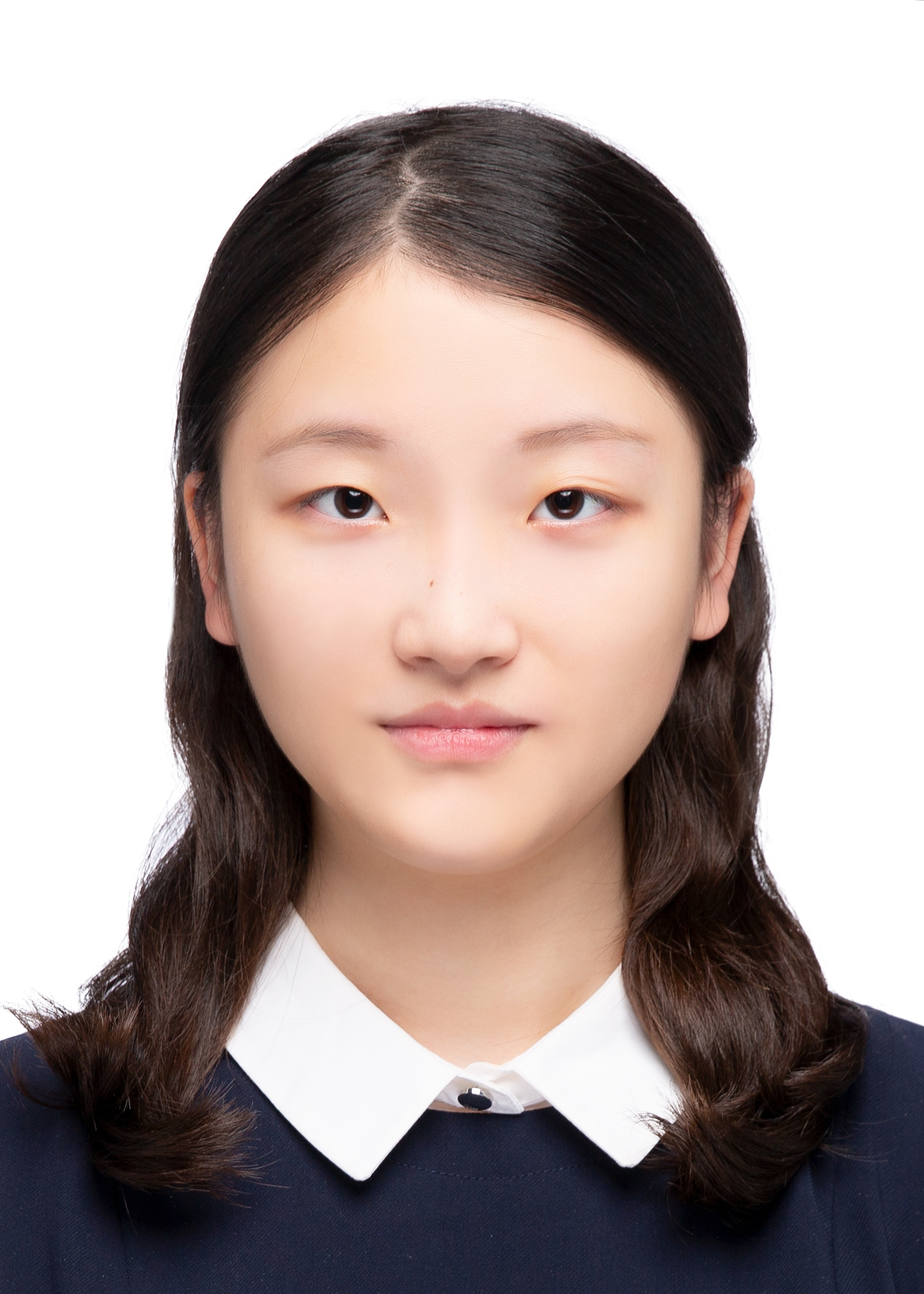}}]{Yuxuan Wang} received the BE degree from Beihang University in 2022. She is working towards the PhD degree in computer software at Tsinghua University. Her research interests include machine learning and time series analysis. 
\end{IEEEbiography}

\vspace{-30pt}
\begin{IEEEbiography}[{\includegraphics[width=1in,height=1.25in,clip,keepaspectratio]{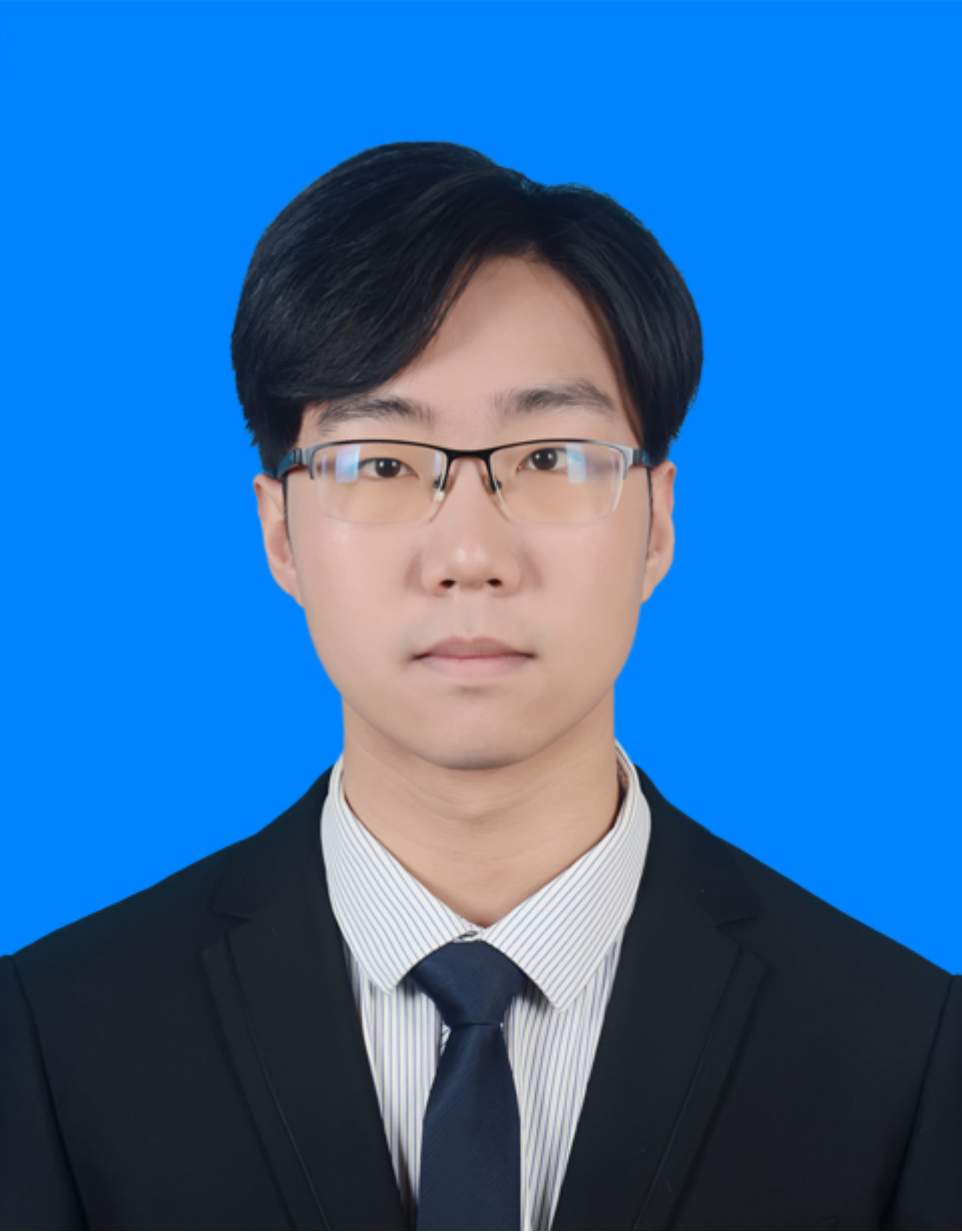}}]{Haixu Wu} received the BE degree in software engineering from Tsinghua University in 2020. He is working towards the PhD degree in computer software at Tsinghua University. His research interests include scientific machine learning, deep learning, and spatiotemporal learning.
\end{IEEEbiography}

\vspace{-30pt}
\begin{IEEEbiography}[{\includegraphics[width=1in,height=1.25in,clip,keepaspectratio]{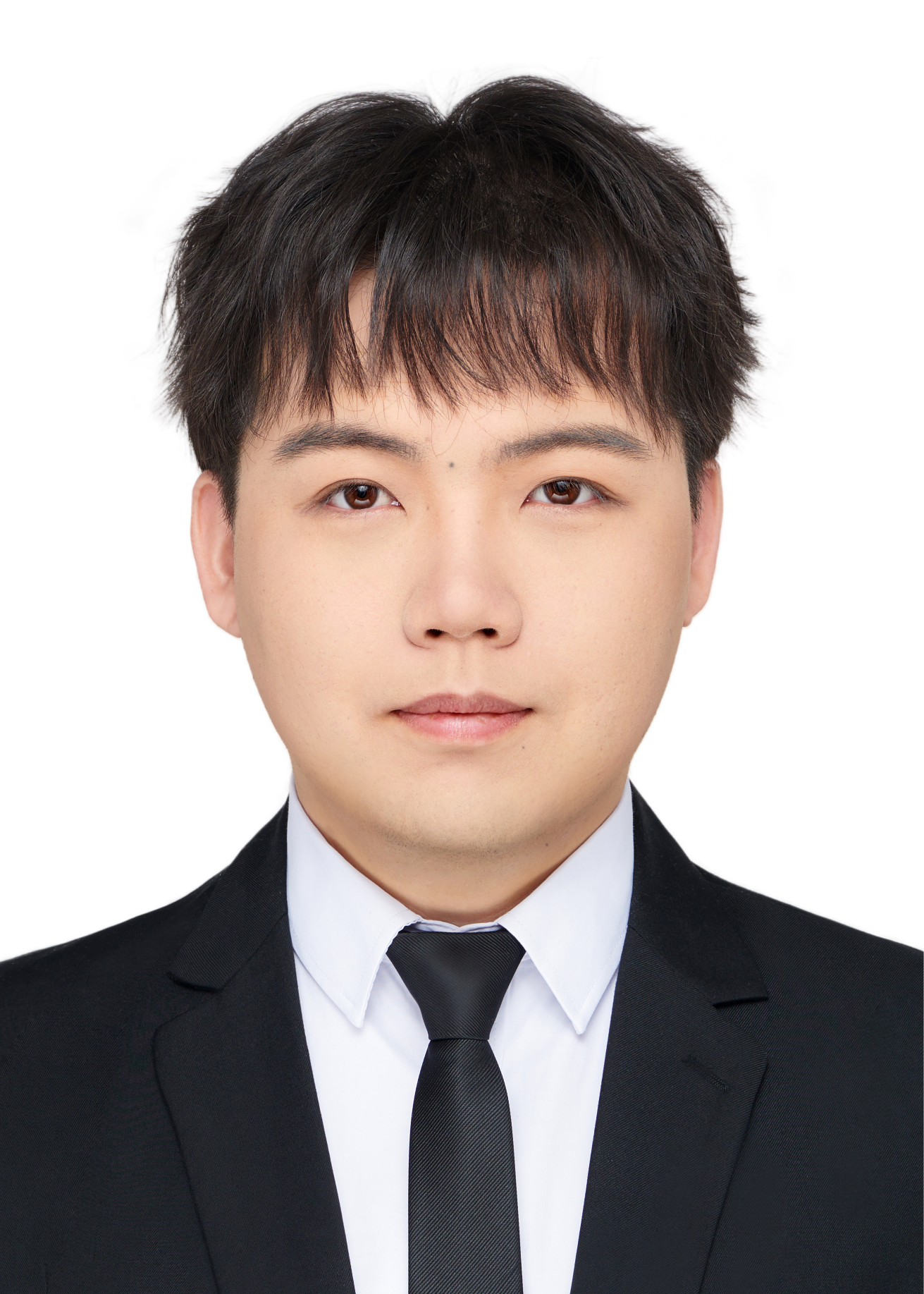}}]{Jiaxiang Dong} received the ME degree in computer science and technology from Nankai University in 2018. He is currently working toward the PhD degree in computer software at Tsinghua University. His research interests include machine learning and time series analysis.
\end{IEEEbiography}

\vspace{-30pt}
\begin{IEEEbiography}[{\includegraphics[width=1in,height=1.25in,clip,keepaspectratio]{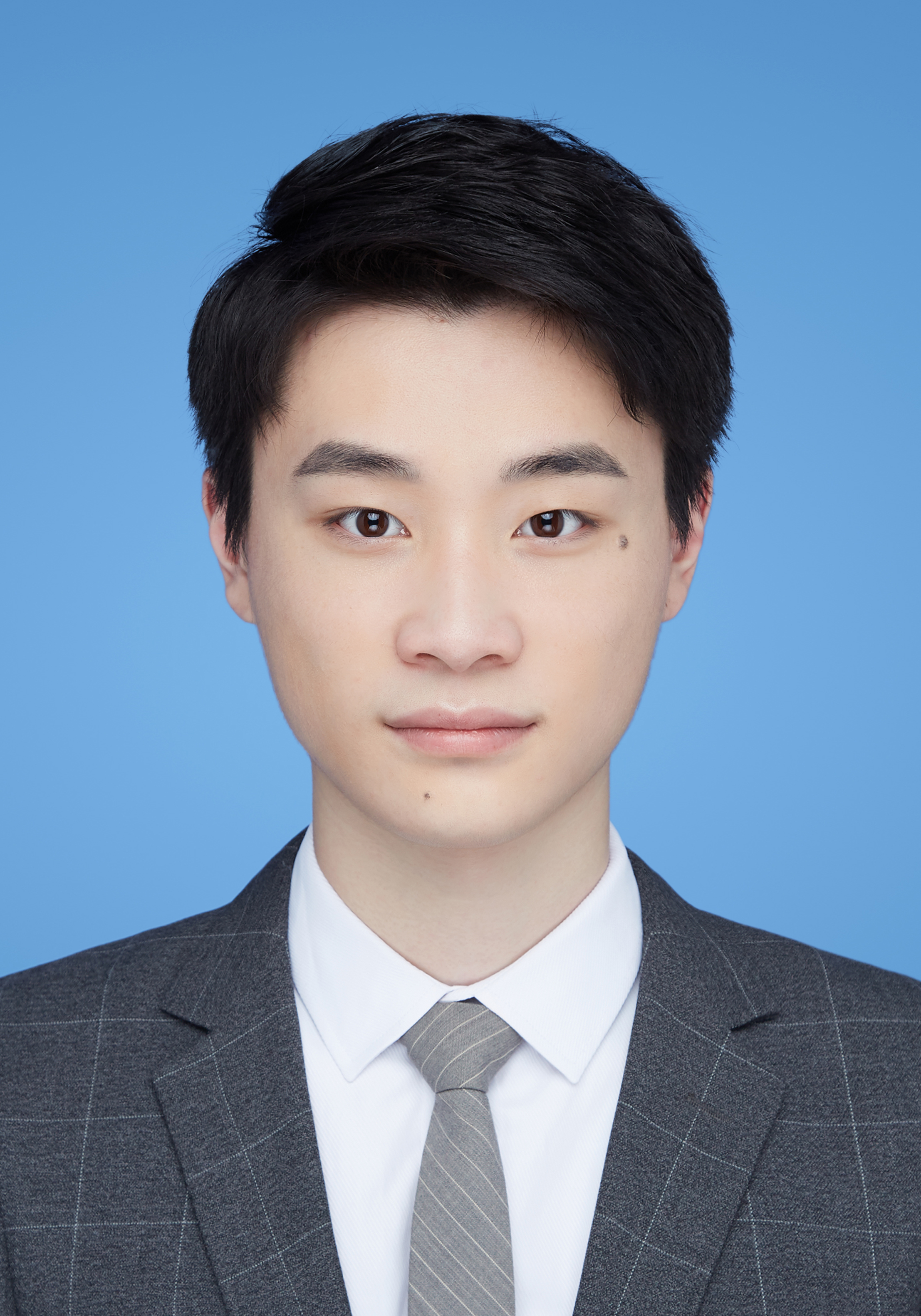}}] {Yong Liu} received the BE degree in software engineering from Tsinghua University in 2021. He is working towards the PhD degree in computer software at Tsinghua University. His research
interests include time series analysis and time series foundation models.
\end{IEEEbiography}

\vspace{-30pt}
\begin{IEEEbiography}[{\includegraphics[width=1in,height=1.25in,clip,keepaspectratio]{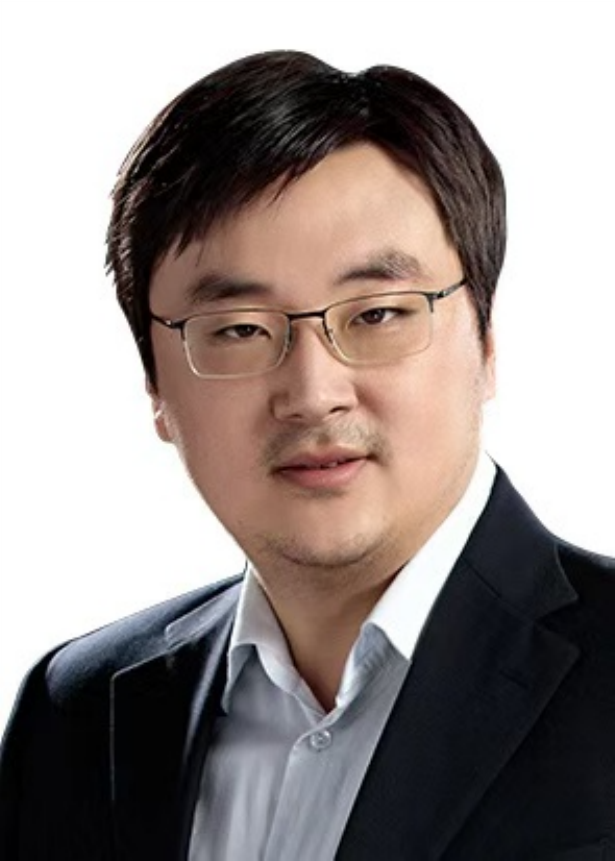}}] {Chen Wang} received B.S. and M.S. degrees in computer science from Fudan University in 2003 and 2006, respectively. He is currently a research associate professor and CTO at National Engineering Laboratory for Big Data Software, Tsinghua University. His research focuses on database systems, data governance, time series data management, and industrial big data applications
\end{IEEEbiography}

\vspace{-30pt}
\begin{IEEEbiography}[{\includegraphics[width=1in,height=1.25in,clip,keepaspectratio]{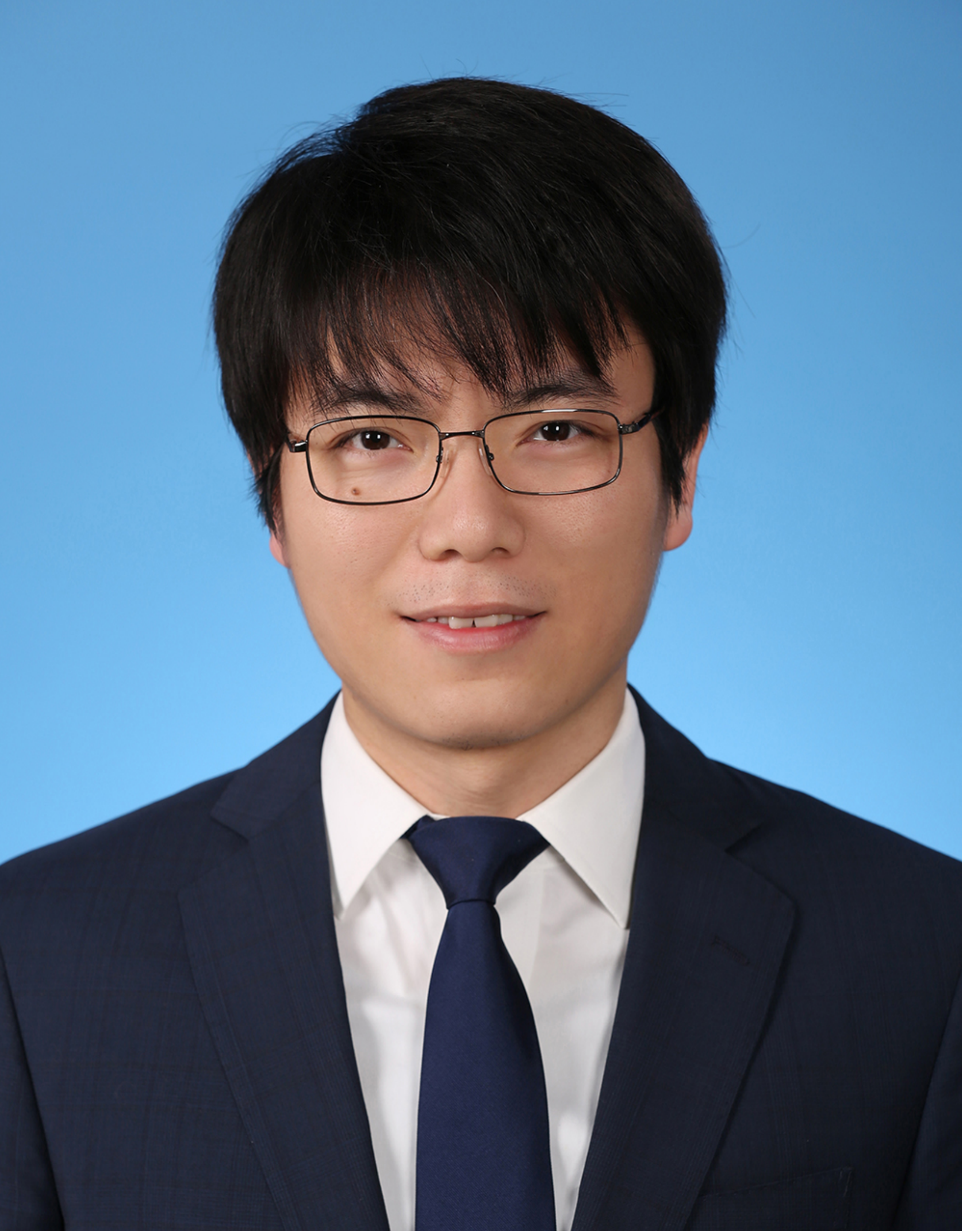}}]{Mingsheng Long} received the BE and PhD degrees from Tsinghua University in 2008 and 2014 respectively. He is a tenured associate professor with the School of Software, Tsinghua University. He serves as an associate editor of \emph{IEEE Transactions on Pattern Analysis and Machine Intelligence} and \emph{Artificial Intelligence Journal}, and as senior/highlighted area chairs of major machine learning conferences, including NeurIPS, ICML, and ICLR. His research is dedicated to machine learning theory, algorithms, and models, with special interests in deep learning and foundation models, scientific learning and world models, transfer learning and domain adaptation.
\end{IEEEbiography}

\vspace{-30pt}
\begin{IEEEbiography}[{\includegraphics[width=1in,height=1.25in,clip,keepaspectratio]{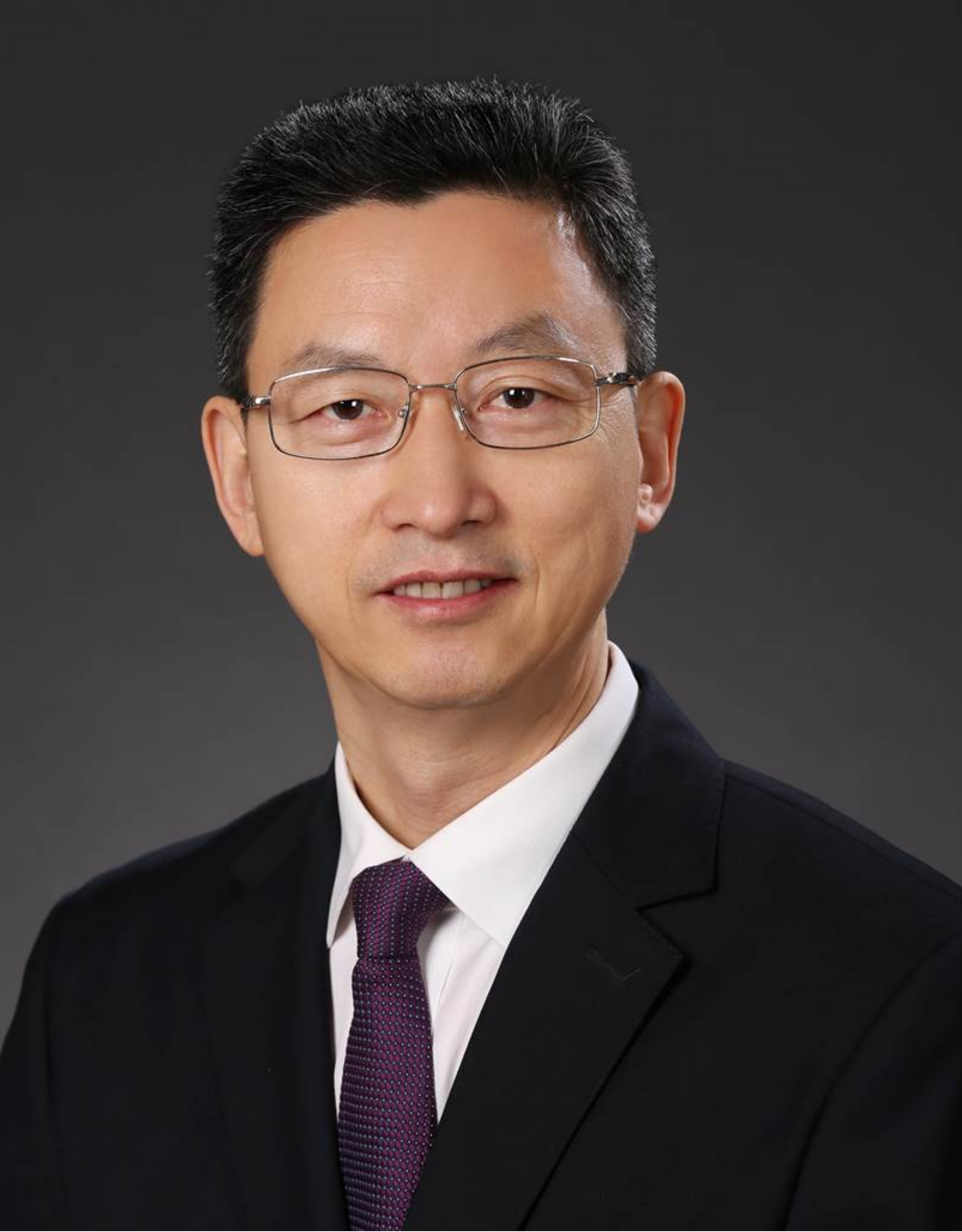}}]{Jianmin Wang} received the BE degree from Peking University in 1990, and the ME and PhD degrees in computer software from Tsinghua University in 1992 and 1995, respectively. He is a full professor with the School of Software, Tsinghua University. His research interests include Big Data management systems and large-scale data analytics. He led to developing a product data and lifecycle management system, which has been deployed in hundreds of enterprises in China. He is leading the development of the Tsinghua DataWay Big Data platform and Apache IoTDB time series database systems in the National Engineering Lab for Big Data Software.
\end{IEEEbiography}

\end{document}